\documentclass[sts]{imsart}

\RequirePackage{amsthm,amsmath,amsfonts,amssymb}
\RequirePackage[numbers]{natbib}
\RequirePackage{color}

\startlocaldefs

\usepackage{shortex}

\usepackage{tikz,pgffor}
\usetikzlibrary{shadows, positioning, arrows, backgrounds, decorations, decorations.pathmorphing, decorations.markings}
\newcommand{\normalLetters}[2]{\letters{#1}{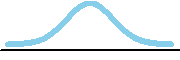}{#2}} 

\newcommand{\letters}[3]{$#1$\\ \includegraphics[scale = 0.25]{lizhen_figure/#2.png} \\ $#3$}

\def\d{\textup{d}}

\def\E{\mathsf{E}}
\def\P{\mathsf{P}}

\makeatletter
\def\greekvectors#1{%
 \@for\next:=#1\do{%
    \def\X##1;{\expandafter\def\csname b##1\endcsname{\bm{\csname##1\endcsname}}}
    \expandafter\X\next;}
 \@for\next:=#1\do{%
    \def\X##1;{\expandafter\def\csname h##1\endcsname{\widehat{\csname##1\endcsname}}}
    \expandafter\X\next;}
 \@for\next:=#1\do{%
    \def\X##1;{\expandafter\def\csname c##1\endcsname{\check{\csname##1\endcsname}}}
    \expandafter\X\next;}
 \@for\next:=#1\do{%
    \def\X##1;{\expandafter\def\csname hb##1\endcsname{\widehat{\bm{\csname##1\endcsname}}}}
    \expandafter\X\next;}
}
\greekvectors{alpha,beta,gamma,delta,epsilon,zeta,theta,kappa,lambda, mu,nu,xi,pi,rho,tau,phi,chi,psi,omega,
    Delta,Gamma,Theta,Lambda,Xi,Pi,Sigma,Phi,Psi,Omega}
    
\@tfor\next:=abcfghijlmnopqrstuwxyzABCDGHIJKLMOQSTUVWXYZ\do{%
    \def\command@factory#1{\expandafter\def\csname #1\endcsname{\mathbf{#1}} }
    \expandafter\command@factory\next
}
\@tfor\next:=abcdefghijklmnpqrstuvwxyzABCDEFGHIJKLMNOPQRSTUVWXYZ\do{%
    \def\command@factory#1{\expandafter\def\csname t#1\endcsname{\widetilde{#1}} }
    \expandafter\command@factory\next
}
\@tfor\next:=abcdefghijklmnopqrstuvwxyzABCDEFGHIJKLMNOPQRSTUVWXYZ\do{%
    \def\command@factory#1{\expandafter\def\csname tb#1\endcsname{\tilde{\mathbf{#1}}} }
    \expandafter\command@factory\next
}
\@tfor\next:=xyzABCDEFGHIJKLMNOPQRSTUVWXYZ\do{%
    \def\command@factory#1{\expandafter\def\csname h#1\endcsname{\widehat{#1}} }
    \expandafter\command@factory\next
}
\@tfor\next:=abcdefghijklmnopqrstuvwxyzABCDEFGHIJKLMNOPQRSTUVWXYZ\do{%
    \def\command@factory#1{\expandafter\def\csname hb#1\endcsname{\widehat{\mathbf{#1}}} }
    \expandafter\command@factory\next
}
\@tfor\next:=abcdefghijklnopqrstuvwxyzABCDEFGHIJKLMNOPQRSTUVWXYZ\do{%
    \def\command@factory#1{\expandafter\def\csname b#1\endcsname{\mathbbm{#1}} }
    \expandafter\command@factory\next
}
\@tfor\next:=ABCDEFGHIJKLMNOPQRSTUVWXYZ\do{%
    \def\command@factory#1{\expandafter\def\csname c#1\endcsname{\mathcal{#1}} }
    \expandafter\command@factory\next
}
\@tfor\next:=abcdefghjklmnopqrstuvwxyzABCDEFGHIJKLMNOPQRSTUVWXYZ\do{
    \def\command@factory#1{\expandafter\def\csname f#1\endcsname{\mathfrak{#1}} }
    \expandafter\command@factory\next
    
}
\@tfor\next:=abcdefghijklmnopqrstuvwxyzABCDEFGHIJKLMNOPQRSTUVWXYZ\do{%
    \def\command@factory#1{\expandafter\def\csname s#1\endcsname{\mathsf{#1}} }
    \expandafter\command@factory\next
}

 \@tfor\next:=abcdefghjklmnopqrstuvwxyzABCDEFGHIJKLMNOPQRSTUVWXYZ\do{
    \def\command@factory#1{\expandafter\def\csname sc#1\endcsname{\mathscr{#1}} }
    \expandafter\command@factory\next
}

\def\bX{\boldsymbol{X}}

\DeclareMathOperator{\PP}{\mathbb{P}} 

\DeclareMathOperator{\RR}{\mathbb{R}} 
\DeclareMathOperator{\Ncal}{\mathcal{N}} 

\endlocaldefs

\begin{document}

\begin{frontmatter}
\title{Machine Learning and the Future of Bayesian Computation}

\begin{aug}
\author[A]{\fnms{Steven}~\snm{Winter}\ead[label=e1]{steven.winter@duke.edu}},
\author[B]{\fnms{Trevor}~\snm{Campbell$^\dag$}\ead[label=e2]{trevor@stat.ubc.ca}},
\author[C]{\fnms{Lizhen}~\snm{Lin$^\dag$}\ead[label=e3]{lizhen.lin@nd.edu}},
\author[D]{\fnms{Sanvesh}~\snm{Srivastava$^\dag$}\ead[label=e4]{sanvesh-srivastava@uiowa.edu}},
\and
\author[E]{\fnms{David B.}~\snm{Dunson}\ead[label=e5]{dunson@duke.edu}}
\address[A]{Steven Winter: PhD Student, Department of Statistical Science, Duke University\printead[presep={\ }]{e1}.}
\address[B]{Trevor Campbell: Assistant Professor, Department of Statistics, University of British Columbia\printead[presep={\ }]{e2}.}
\address[C]{Lizhen Lin: Robert and Sara Lumpkins Associate Professor, Department of Applied and Computational Mathematics and Statistics, University of Notre Dame\printead[presep={\ }]{e3}.}
\address[D]{Sanvesh Srivastava: Associate Professor, Department of Statistics and Actuarial Science, University of Iowa\printead[presep={\ }]{e4}.}
\address[E]{David B. Dunson: Arts and Sciences Distinguished Professor, Departments of Statistical Science and Mathematcs, Duke University\printead[presep={\ }]{e5}.}
\end{aug}

\begin{abstract}
Bayesian models are a powerful tool for studying complex data, allowing the analyst to encode rich hierarchical dependencies and leverage prior information. Most importantly, they facilitate a complete characterization of uncertainty through the posterior distribution. Practical posterior computation is commonly performed via MCMC, which can be computationally infeasible for high dimensional models with many observations. In this article we discuss the potential to improve posterior computation using ideas from machine learning. Concrete future directions are explored in vignettes on normalizing flows, Bayesian coresets, distributed Bayesian inference, and variational inference.
\end{abstract}


\end{frontmatter}

\def\thefootnote{$\dag$}\footnotetext{These authors contributed equally.}\def\thefootnote{\arabic{footnote}}

\section{Introduction}
There is immense interest in performing inference and prediction for complicated real-world processes within science, industry, and policy. Bayesian models are appealing because they allow specification of rich generative models encompassing hierarchical structures in the data, natural inclusion of information from experts and/or previous research via priors, and a complete characterization of uncertainty in learning/inference/prediction through posterior and predictive distributions. The primary hurdle in applying Bayesian statistics to complex real-world data is posterior computation. In practice, posterior computation -- evaluating posterior probabilities/expectations, credible intervals for parameters, posterior inclusion probabilities for features, posterior predictive intervals, etc -- 
is typically based on posterior samples using Markov chain Monte Carlo (MCMC). Standard MCMC approaches often fail to converge when the posterior has complicated geometry, such as multiple distant modes or geometric/manifold constraints. Even sampling from simple posteriors can be challenging when the data has tens or hundreds of millions of observations. This article focuses on the future of Bayesian computation, with emphasis on posterior inference for high dimensional, geometrically complicated targets with potentially millions of datapoints.

The recent explosive success of machine learning is key in shaping our vision for the future of Bayesian computation. To make our vision concrete, we have prepared four vignettes covering disjoint cutting-edge computational techniques, all involving ideas from machine learning. The first vignette describes normalizing flows as a new tool for adaptive MCMC with complicated targets; the second describes Bayesian coresets as a method of data compression prior to sampling; the third describes distributed Bayesian inference for huge datasets; the fourth describes modern variational inference for settings where the previous techniques falter. All sections focus heavily on promising avenues for future research.

\section{Sampling using Deep Generative Models}
The Metropolis Hastings (MH) algorithm (often within Gibbs) is by far the most popular tool for sampling posterior distributions \citep{dunson2020hastings}. Good mixing is critically dependent on how closely the MH proposal distribution mimics the target distribution. Higher dimensional targets with increasingly complicated geometry require increasingly flexible proposal distributions which become difficult to tune. Consequently, it is routine to settle for simpler proposals which provide a good local approximation to the target, such as a multivariate Gaussian. Parameters are then tuned to encourage efficient exploration, e.g. by adaptively learning the posterior covariance \citep{haario2001adaptive, roberts2009examples, vihola2012robust} or by discretizing dynamics driven by the target \citep{neal2011mcmc, ma2015complete}. A major limitation of local methods is their practical inability to cross low-probability regions, resulting in poor convergence rates for multimodal distributions \citep{mangoubi2018does}. Many solutions have been proposed, ranging from slightly modifying local kernels to encourage crossing low probability regions \citep{lan2014wormhole, nishimura2016geometrically, lu2017relativistic} to constructing entirely new kernels which are mixtures of a local and global component \citep{andricioaei2001smart, ahn2013distributed, pompe2020framework}. Despite these advances, there is still no general solution for efficiently sampling complicated high dimensional distributions. 

We believe deep learning will play an integral role in developing better general solutions. Deep generative models have demonstrated remarkable success in estimating and approximately sampling complicated, high dimensional distributions, achieving state-of-the-art performance in image/audio/video synthesis, computer graphics, physical/engineering simulations, drug discovery, and other domains \citep{harshvardhan2020comprehensive, kobyzev2020normalizing}. In this vignette we discuss the use of deep generative models to design better proposal distributions for use in MH, both by augmenting existing kernels and by constructing entirely new distributions. Most deep generative models use a neural network (NN) to transform a simple base distribution to closely match a pre-specified empirical distribution. The setting of posterior computation via MH introduces two practical problems. First, samples from the target are not available prior to sampling, complicating the process of training the NN. Second, each iteration of MH requires computing the acceptance probability, hence evaluating the proposal density. If the proposal is a simple distribution transformed by a NN, then this requires inverting a NN, which is generally impossible, and computing the Jacobian, which can be numerically intractable in high dimensions.

In this vignette we discuss adaptively tuning normalizing flow (NF) proposals as a means of resolving these challenges. Section 2.1 introduces NFs; Sections 2.2-2.3 cover applications to MH and straightforward generalizations; Section 2.4 discusses exciting future research.

\subsection{Introduction to normalizing flows}
In this section we provide a brief introduction to NFs and highlight useful properties. One method for generating a flexible class of proposal distributions is to transform a simple $D$-dimensional random variable $Z$ (e.g., $Z\sim N(0, I_D)$) with a diffeomorphism $f$ parameterized by a NN. Carefully tuning $f$ can result in proposals $Y=f(Z)$ that closely conform to the target. Computing the acceptance probability in each iteration of MH requires evaluating the proposal density,
\begin{equation}
    \pi_Y(y) = \pi_Z(f^{-1}(y))|J_{f^{-1}}(f^{-1}(y))|\label{eq:change_vars} 
\end{equation}
where $\pi_Z$ is the density of $Z$ and $J_{f^{-1}}$ is the Jacobian of $f^{-1}$. Inverting NNs is generally intractable, and evaluating Jacobians is $O(D^3)$ in the worst case. 

NFs impose additional structure on $f$ to resolve these problems. Specifically, discrete NFs (DNFs) decompose $f$ as the composition of $K$ simple component functions:
\begin{equation}
    f = f_K \circ \cdots \circ f_1.
\end{equation}
Component functions are constructed to facilitate fast inversion (either exactly or approximately) and fast Jacobian calculations (e.g., by  ensuring Jacobians are upper/lower triangular). The change of variables rule becomes
\begin{equation}
    \pi_Y(y) = \pi_Z(f^{-1}(y))\prod_{i=1}^K|J_{f_i^{-1}}(z_i)|
\end{equation}
where $f^{-1}=f_1^{-1} \circ \cdots \circ f_K^{-1}$ and $z_i=f_{i+1}^{-1} \circ \cdots \circ f_K^{-1}(y)$ with $z_K=y$. By the inverse function theorem, $J_{f_i^{-1}}=J_{f_i}^{-1}$, so it is sufficient to compute the Jacobian of $f_i$ or $f_i^{-1}$. For example, a \textit{planar} normalizing flow \citep{rezende2015variational} uses component functions
\begin{equation}
    f_i(z) = z + a_ih(w_i^Tz+b_i)\label{eq:planar}
\end{equation}
where $a_i,w_i\in\mathbb{R}^D$, $b_i\in\mathbb{R}$ are parameters to be tuned and $h$ is an invertible, differentiable nonlinearity applied elementwise. The matrix determinant lemma allows one to express the Jacobian as
\begin{equation}
    |J_{f_i}(z)| = 1+h'(w_i^Tz+b)a_i^Tw_i
\end{equation}
which is $O(D)$ to compute. Planar flows are not invertible for all choices of parameters and nonlinearities, however efficient constrained optimization algorithms are available which ensure invertibility \citep{rezende2015variational}. Planar flows have relatively limited expressivity, and many layers may be needed to construct suitably complicated high dimensional proposals. Improved component functions have been proposed, including radial \citep{rezende2015variational}, spline \citep{durkan2019neural}, coupling \citep{dinh2016density}, autoregressive \citep{kingma2016improved}, etc. See \citep{kobyzev2020normalizing} for a review of NFs and \citep{papamakarios2021normalizing} for theory on the expressively of discrete flows. 

Continuous normalizing flows (CNFs) \citep{chen2018neural} are an extension of the discrete framework, potentially enhancing expressivity while requiring fewer parameters and lower memory complexity. The key insight is to reconceptualize DNFs as a method for computing the path $x(t)$ of a particle at discrete times $t\in\{0,1/K,2/K...,1\}$. The initial location $x(0)$ is drawn from $Z$. At time $1/K$, the location is updated to $x(1/K)=f_1(x(0))$. This is repeated iteratively, moving from $x(i/K)$ at time $i/K$ to $x((i+1)/K)=f_i(x(i/K))$ at time $i+1$. The result is a path $(x(0),...,x(1))$ where the final location is a sample from $Y$. CNFs consider the limit $K\to\infty$, with the intuition that one can obtain a more flexible distribution for $Y$ by flowing samples of $Z$ through continuous paths instead of discrete paths. This can be formalized as the initial value problem
\begin{equation}
    \frac{dx(t)}{dt} = f(x(t), t) \label{eq:ODE}
\end{equation}
where $f$ is a function parameterized by a NN and $x(0)$ is a sample from $Z$. In practice equation \eqref{eq:ODE} cannot be solved analytically, however approximate samples of $Y$ can be generated using an ODE solver. Euler's method with a step size of $1/K$ exactly recovers a DNF with $K$ layers, but greater expressivity can be obtained using higher order solvers. This framework has a number of surprising technical advantages; see \citep{chen2018neural} for an exposition.

\subsection{Normalizing flow proposals}
In this section we outline modern methods for constructing proposals with NFs. Throughout, we denote the $D$-dimensional target density by
\begin{equation}
    \pi(x) \propto \exp(-U(x))
\end{equation}
with unknown normalizing constant and known potential $U:\mathbb{R}^{D}\to \mathbb{R}$. A NF with parameters $\phi$ will be denoted $f_\phi:\mathbb{R}^D\to\mathbb{R}^D$; this yields a new density $\hat\pi_\phi$ by pushing forward a simple random variable $Z$ with density $\pi_Z$.

\paragraph*{\textbf{Independent proposals}}
The simplest approach is to use a NF to generate proposals in independent MH \citep{brofos2022adaptation}. At each iteration, a proposed state $x'$ is generated by pushing a sample of $Z$ through the NF. This state is accepted with probability
\begin{equation}
    \text{acc}(x,x')=\min\bigg\{1,\frac{\pi(x')\hat\pi_\phi(x)}{\pi(x)\hat\pi_\phi(x')}\bigg\}\label{eq:MH}
\end{equation}
where $x$ is the current state. In high dimensions, almost all choices of $\phi$ will result in low overlap between $\hat\pi_\phi$ and $\pi$, hence small acceptance ratios and poor mixing. Consequently, we focus our discussion on more elaborate proposals which result in better practical performance.

\paragraph*{\textbf{Dependent proposals}}
A more practical approach is to let proposals depend on the current state. This can be achieved by using a larger NF $f_\phi:\mathbb{R}^D\times \mathbb{R}^M\to \mathbb{R}^D\times \mathbb{R}^M$ which maps the current state $x$ and $M$-dimensional noise $z$ to a proposal $x'$ and transformed noise $z'$. The $M$ dimensional noise can be thought of as an auxillary parameter such as momentum or temperature in dynamics based MCMC. \citep{song2017nice} construct a dependent proposal which is symmetric, thus eliminating the ratio of proposal densities in equation \eqref{eq:MH} and reducing the problem of extremely low early acceptance rates. The proposal is constructed in two stages: first, sample $u\sim \text{Uniform}[0,1]$ and $z$ from $Z$. If $u>0.5$, propose $x'$ using $(x',z')=f_\phi(x,z)$. Otherwise propose $x'$ using $(x',z')=f_\phi^{-1}(x,z)$. Using a mixture of $f_\phi$ and $f_\phi^{-1}$ ensures that $x'$ is as likely to be proposed when starting at $x$ as $x$ is to be proposed when starting at $x'$. Key to the proof of symmetry is the assumption that the NF is volume preserving. This is a restrictive assumption: current volume preserving architectures are outperformed by non-volume preserving architectures.

\paragraph*{\textbf{Mixture kernels}}
Higher initial acceptance rates can be obtained by combining NF proposals with classical kernels, for example by alternating proposing samples with HMC and a conditional flow. Samples from the classical kernel provide data with which to tune the NF. Eventually, the NF becomes a good approximation to the posterior, proposing efficient global moves and resulting in better mixing than the classical kernel alone. \citep{gabrie2022adaptive} construct a proposal which deterministically alternates between approximately $10$ MALA proposals for every one independent NF proposal. The resulting sampler efficiently explores multimodal distributions: MALA locally explores each mode, and NF teleports the chain between modes. It is critical to initialize the sampler with at least one particle in each mode, as the local dynamics are unlikely to discover new modes on their own. The algorithm is shown to converge with an exponential rate in the continuous time limit. Partial ergodic theory is available when the flow is adaptively learned by minimizing the KL divergence, although other loss functions remain unstudied.

\paragraph*{\textbf{Augmenting existing kernels}}
The previously discussed mixtures rely on classical kernels for local exploration until there is sufficient data to train the NF. An alternate approach is to use NFs to augment classical kernels - that is, to improve the classical kernel as the chain runs instead of tuning a separate, auxillary kernel. We use HMC as an example, wherein a new state $x'$ is proposed by drawing a momentum $\nu\sim N(0, I_D)$ and approximating the resulting Hamiltonian dynamics (usually) with the leapfrog integrator. One time step of the approximation proceeds by taking a half step of the momentum
\begin{equation}
\nu_{1/2} = \nu-\frac{\varepsilon}{2}\nabla U(x)
\end{equation}
where $x$ is the current state and $\varepsilon$ is the step size of the integrator. This is used to update the position
\begin{align}
    x' = x + \varepsilon\nu_{1/2}
\end{align}
which is then used to update the momentum,
\begin{equation}
\nu' = \nu-\frac{\varepsilon}{2}\nabla U(x')
\end{equation}
The process is repeated a prespecified number of times to generate a final proposal; the final momentum is disregarded. The resulting proposal is symmetric and volume preserving, resulting in a simple acceptance ratio. Crossing low-probability regions requires a large velocity, which is unlikely if the momentum is sampled from a Gaussian. \citep{levy2017generalizing} use NFs to learn a collection of maps which dynamically rescale the momentum and position to encourage exploration across low probability regions. Specifically, the momentum half step is replaced by
\begin{equation}
\nu_{1/2} = \exp(S_{\nu}(x))\odot\nu-\frac{\varepsilon}{2}\exp(Q_{\nu}(x))\odot\nabla U(x)+T_{\nu}(x)\label{eq:L2HMC}
\end{equation}
where $\odot$ is the elementwise product, $S_\nu$ is a NF that rescales the momentum, $Q_\nu$ is a NF that rescales the gradient, and $T_\nu$ is a NF that translates the momentum. Similarly, the position update is replaced with 
\begin{equation}
    x' =  \exp(S_{x}(\nu_{1/2}))\odot x + \varepsilon\exp(Q_{x}(\nu_{1/2}))\odot\nu_{1/2}+T_{x}(\nu_{1/2})
\end{equation}
where $S_x$, $Q_x$, and $T_x$ are NFs. The momentum is updated again with equation \eqref{eq:L2HMC} using $x'$ in place of $x$, and the entire procedure is iterated. When all of these NFs are zero, we exactly recover HMC. Allowing the NFs to be nonzero results in a very flexible family of proposal distributions which can be adaptively tuned to propel the sampler out of low probability regions by rescaling and translating the momentum/position. The invertibility and tractable Jacobians allows efficient calculation of the proposal density. This presentation has been simplified from \citep{levy2017generalizing}, which also includes random directions, random masking, and conditions NFs on the leapfrog iteration. So far, the above augmentation technique has only been applied to HMC. However there is a broad class of dynamical systems that can be used to generate proposals, including Langevin dynamics, relativistic dynamics, Nose-Hoover thermostats, and others \citep{ma2015complete}. NFs can be used to augment all of these algorithms using the same recipe as above.

\subsection{Tuning proposals}
Appropriately tuning NF parameters is critical for good mixing. In practice, tuning is often performed by adaptively minimizing a loss. In this section we cover a variety of candidate loss functions, including measure-theoretic losses, summary statistics, and adversarial approaches.

\paragraph*{\textbf{Statistical deviance}}
The simplest approach is to define a function $d$ measuring how close the proposal is to the target and then to find NF parameters minimizing $d(\hat\pi_\phi, \pi)$. Let $\mathcal{G}$ be a space of probability densities and $d:\mathcal{G}\times \mathcal{G}\to\mathbb{R}$ be any function measuring the distance/discrepancy/deviance between two probability measures. We assume 
\begin{enumerate}
    \item $d(\rho,\rho)=0$ for all $\rho\in\mathcal{G}$.
    \item $d(\rho,\rho')>0$ for $\rho\neq \rho'\in\mathcal{G}$, with equality interpreted as equality almost everywhere.
    \item $d(\hat\pi_\phi, \pi)$ has a gradient with respect to $\phi$, $\nabla_\phi d(\pi_\phi, \pi)$.
\end{enumerate}
Conditions $(1)$ and $(2)$ ensure $d(\hat\pi_\phi, \pi)=0$ if and only if $\hat\pi_\phi=\pi$, hence minimizing $d$ is a reasonable way to approximate the target. Condition $(3)$ allows optimization with gradient based methods. Weaker notions of differentiablility are sufficient, such as having a tractable subgradient.

For example, $d$ may be the forward KL divergence,
\begin{equation}
    \mathrm{D}_{\mathrm{KL}}\left(\pi \| \hat\pi_\phi\right)= \int_{\mathbb{R}^D} \pi(x)\log\bigg(\frac{\pi(x)}{\hat \pi_\phi(x)}\bigg)dx \label{eq:KL}
\end{equation}
Adaptive estimation can be performed by alternating between generating a sample via MH and updating NF parameters using the gradient of \eqref{eq:KL} \citep{brofos2022adaptation}. The gradient can be estimated via Monte Carlo using previous samples. Under technical assumptions on the NF and the target, the resulting Markov chain is ergodic with the correct limiting distribution \citep{brofos2022adaptation}. 

Other viable choices for $d$ include the Hellinger distance, the (sliced) Wasserstein distance, the total variation distance, etc. Many of these are as-of-yet unexplored as a means of adaptively estimating flows, and it is unclear which will result in the best performance. The main limitation with approaches in this class is that minimizing a difference only indirectly targets good mixing; in the following we consider directly targeting good mixing with MCMC diagnostics.

\paragraph*{\textbf{Mixing summary statistics}}
A high quality global approximation of the target may not be required for sufficiently good mixing, especially if NFs are used in conjunction with local kernels such as HMC. Using distance based losses in these situations is unnecessarily ambitious and better practical performance may be attained by switching to a loss function which directly targets good mixing. Ideally one would maximize the effective sample size, but this depends on the entire history of the chain and is in general slow to compute. Instead \citep{levy2017generalizing} propose minimizing the lag-1 autocorrelation, which is equivalent to maximizing the expected squared jump distance \citep{pasarica2010adaptively}:
\begin{equation}
    \text{lag}(\hat\pi_\phi, \pi) = E[||x-x'||_2^2\text{acc}(x,x')]
\end{equation}
where the expectation is over the target and any auxiliary variables used to sample $x'$. This can be estimated using samples $x_i$, $i=1,...,S$ from the first $S$ iterations of the chain by generating a proposal $x_i'$ starting at each $x_i$ and averaging:
\begin{equation}
\text{lag}(\hat\pi_\phi, \pi) \approx \frac{1}{S}\sum_{i=1}^S||x_i-x_i'||_2^2\text{acc}(x_i,x_i').
\end{equation}
This loss depends on $\phi$ implicitly through the $x_i'$. Naively optimizing this loss does not guarantee good mixing across the entire space - for example, the chain may bounce between two distant modes. To solve these problems, \citep{levy2017generalizing} add a reciprocal term and instead optimize
\begin{equation}
\ell_\lambda(\hat\pi_\phi, \pi) = \frac{\lambda}{ \text{lag}(\hat\pi_\phi, \pi)} - \frac{\text{lag}(\hat\pi_\phi, \pi)}{\lambda}
\end{equation}
where $\lambda>0$ is a tuning parameter. The reciprocal term penalizes states where the expected squared jump distance is small. \citep{levy2017generalizing} add a term of the same form to encourage faster burn-in. The composite loss is used to train an augmented variant of HMC and results in a sampler which efficiently moves between well-separated modes.

Other summary statistics can be integrated into this framework, possibly considering lag-$k$ autocorrelations or multiple chain summaries such as the Gelman-Rubin statistic \citep{gelman1992inference}. One concern with this class of loss functions is that no single summary statistic can detect when a chain has mixed, and naively optimizing one statistic may result in pathological behaviour that is hard to detect. In the following we discuss a different strategy which may strike a middle ground between ambitious distance based methods and narrow summary statistic based methods.

\paragraph*{\textbf{Adversarial training}}
Generative adversarial networks (GANs) \citep{goodfellow2020generative, gui2021review} pit two NNs against each other in a minimax game. The first player is a generator which transforms noise into samples that look like real data; the second player is a discriminator which tries to determine whether an arbitrary sample is synthetic or real. GANs may be applied to MCMC by taking the proposal distribution to be the generator and training a discriminator to distinguish between proposals and previous samples of the target. \citep{song2017nice} use this idea to adaptively train a NF proposal which dramatically outperforms HMC on multimodal distributions. 

Many improvements are possible by leveraging modern ideas from the GAN literature. Conditional GANs \citep{mirza2014conditional} allow the discriminator and generator to condition on external variables. For example, one could construct a tempered adversarial algorithm by conditioning on a temperature variable, possibly accelerating the mixing of annealed MCMC. Complicated GAN structures are prone to mode collapse, hence these generalizations will likely require modified loss functions \citep{uehara2016generative, mao2017least, jolicoeur2018relativistic, wang2018improving} and regularization \citep{gulrajani2017improved, petzka2017regularization, roth2017stabilizing, miyato2018spectral}.

\subsection{Future directions}
We have introduced several different kernel structures and losses which can be combined to develop new adaptive MCMC algorithms. In this section we discuss shortcomings of the proposed approach, as well as avenues for exciting long-term research.

\paragraph*{\textbf{Theoretical guarantees}}
So far, partial ergodic theory is only available in the simplest case of tuning an independent NF proposal by adaptively minimizing the KL divergence \citep{brofos2022adaptation}. Dependent/conditional proposals and augmented kernels are not well studied, and no guarantees are available when adaptively minimizing summary statistic or adversarial based losses. This is particularly concerning for summary statistic based losses, as it is not clear that minimizing (e.g., lag-1 autocorrelation) is enough to guarantee ergodic averages converge to the correct values. Precise theoretical results will provide insights into when/why these methods succeed/fail, and are a necessary precursor to widespread adoption of NF sampling.

\paragraph*{\textbf{Constrained posteriors}}
In this vignette we only consider the case where the target is supported over Euclidean space, however in some applications the target is supported over a Riemannian manifold (e.g., the sphere or positive semidefinite matrices). Most manifold sampling algorithms rely on approximating dynamics defined either intrinsically on the manifold or induced by projecting from ambient space. These dynamics based methods may be inferior to NF kernels for multimodal distributions. Recent work has successfully generalized NFs to Riemannian manifolds, although these constructions typically place significant restrictions on geometry (e.g., diffeomorphic to a cross product of spheres \citep{rezende2020normalizing}) or rely on high-variance estimates of Jacobian terms \citep{mathieu2020riemannian}. Loss functions measuring the distance between a proposal and the target may be harder to define and compute over manifolds. New architectures for manifold valued NFs and improved estimation techniques could facilitate efficient sampling in a wide class of models with non-Euclidean supports.

Our discussion also neglected to mention discrete parameters. Discrete parameters occur routinely in Bayesian applications, including clustering/discrete mixture models, latent class models, and variable selection. Specific NF architectures have been constructed to handle discrete data \citep{tran2019discrete, ziegler2019latent}, but current approaches are relatively inflexible and cannot be made more flexible by naively adding more NN layers, limiting their utility within MH. A more promising direction is to leverage the flexibility of continuous NFs by embedding discrete parameters in Euclidean space and sampling from an augmented posterior. Several variants of HMC have been proposed to accommodate piecewise discontinuous potential functions \citep{pakman2013auxiliary, mohasel2015reflection, dinh2017probabilistic}, with recent implementations such as discontinuous HMC (DHMC) \citep{nishimura2017discontinuous} achieving excellent practical performance sampling ordinal variables. However, embedding based methods struggle to sample unordered variables - here the embedding order is arbitrary, with most embeddings introducing multimodality in the augmented posterior. NFs have successfully augmented continuous HMC \citep{levy2017generalizing} to handle multimodal distributions; the same strategy is promising for improving DHMC. 

\paragraph*{\textbf{Automated proposal selection}}
{\em A priori} it is unclear which NF architecture, kernel structure, and loss will result in the most efficient mixing for sampling a given posterior. Running many Markov chains with different choices can be time consuming, and a large amount of computational effort may be wasted if some chains mix poorly. Tools for automatic architecture/kernel/loss selection would greatly improve the accessibility of the proposed methodology. This goal is difficult in general given (1) the space of possible samplers is huge, (2) different architectures and kernels are not always comparable, and (3) good mixing is impossible to quantify with a single numerical summary. 

Ideas from reinforcement learning, sequential decision making, and control theory could provide principled algorithms for exploring the space of possible samplers. One could define a state space of kernel/loss pairs, $(\hat\pi_\phi, L)$, which an agent interacts with by running adaptive MCMC. After each action, the agent observes sampler outputs such as trace plots and summary statistics. The goal is to develop a policy for choosing the next kernel/loss pair to run while maximizing some cumulative reward, such as cumulative effective sample sizes across all chains. As an initial attempt, one could restrict kernels to all have the same structure, such as HMC/NF mixtures where only the NF architecture is changing, and the loss function to be a simple parametric family, such as the lag-1 loss with different tuning parameters. This facilitates a parameterization of the state-space and allows application of existing continuous-armed bandit algorithms \citep{agrawal1995continuum, wang2008algorithms}. Constructing a sequential decision making algorithm that can efficiently explore kernel/loss pairs with fundamentally different kernel structures and loss functions is an open challenge, which will likely require better understanding of the theoretical relationships between the different proposed kernel structures as well as the dynamics which result from minimizing different types of losses. 

We expect broad patterns to emerge with increasing use of NF, with certain architectures/kernels/losses performing consistently well in specific classes of problems. For example, the authors have observed that discrete spline flows work very well for sampling from Gaussian mixture models. These heuristics could be collected in a community reference manual, allowing statisticians to quickly find promising candidate algorithms for their model class, dimension, features of the data, etc. Crowd-sourcing the construction and maintenance of this manual could enable statisticians to stay up-to-date with NFs, despite the rapid pace of ML research.

\paragraph*{\textbf{Accelerated tuning}}
The recipe presented in this vignette is to (1) choose a NF kernel structure, (2) choose a loss, and (3) adaptively estimate parameters starting from a random initialization. Starting from a random initialization in step (3) is inefficient. Transfer/meta learning may provide tools for accelerating tuning by avoiding random initialization. For example, iterative model development and sensitivity analysis often involve repeating the same inferences with slightly different prior specifications. NF parameters estimated for one prior specification could be used to initialize the sampler for other prior specifications, potentially eliminating the need for adaptive tuning.

A more difficult task is handling targets with similar structures, but different dimensions. For example, consider a Bayesian sparse logistic regression model classifying Alzheimer's disease status using vectorized images of brains. Interest is in sampling coefficients $\beta_I$ from the posterior $\pi(\beta_I\mid A, I)$ where $A=(A_1,...,A_n)$ is a set of disease indicators and $I=(I_1,...,I_n)$ is a set of brain images. Perhaps additional covariates for each subject are collected at a later stage, such as gene expression vectors $G=(G_1,...,G_n)$. Intuitively, there should be strong similarities between the updated posterior $\pi(\beta_I, \beta_G\mid A, I,G)$ and the original posterior $\pi(\beta_I\mid A, I)$, however this is difficult to formalize because the posteriors have different dimensions. 

A promising approach is to parameterize the initial sampler in a dimension-free manner, for example by defining a kernel which proposes an update for the $i$th coefficient depending only on the potential $U(\beta)$, the gradient in that direction $\partial_{\beta_i}U(\beta)$, and auxillary variables in that direction. This kernel can be tuned while sampling $\pi(\beta_I\mid A, I)$ with any of the aforementioned loss functions, and then automatically applied to sample $\pi(\beta_I, \beta_G\mid A, I,G)$. \citep{gong2018meta} introduce a related idea for stochastic gradient sampling of Bayesian neural networks with different activation functions. The general methodology remains unstudied for exact sampling. The proposed coordinate-wise strategy cannot leverage correlation between pairs of parameters to propose efficient block updates; solutions to this problem constitute ongoing research.

\section{Bayesian Coresets}
Large-scale datasets---i.e., those where even a single pass over the complete
data set is computationally costly---are now commonplace. MCMC typically requires
many passes over the full data set; in the setting of large-scale data, this
makes inference, iterative model development, tuning, and verification arduous
and error-prone.  To realize the full benefits of Bayesian
methods in important modern applications, we need inference algorithms that handle the
scale of modern datasets. 

In the past decade, there has been a flurry of work on approximate Bayesian 
inference methods that are computationally efficient in the large-scale data regime.
One class of methods---including variational inference
\cite{Jordan99,Wainwright08,Blei17} and Laplace approximations
\cite{shun1995laplace,hall2011asymptotic}---formulates inference as an
optimization problem that can be solved via (scalable) stochastic gradient
descent \cite{Hoffman13,Ranganath14}.  Because the problem is generally
nonconvex, these approaches come with little or no realizable guarantees, 
and tend to be
sensitive to initialization, optimization hyperparameters, and stochasticity
during optimization.
Another class---subsampling MCMC \cite{Bardenet15,Korattikara14,Maclaurin14,Welling11,Ahn12};
see \citet{Quiroz18} for a recent survey---run a Markov chain whose transitions depend on a subset
of data randomly chosen at each iteration. However, speed benefits can be outweighed by drawbacks, 
as uniformly subsampling at each step causes MCMC to either mix slowly or provide poor
approximation \cite{Johndrow20,Nagapetyan17,Betancourt15,Quiroz18,Quiroz19}.
It is possible to circumvent this restriction by design of an effective control variate for the log-likelihood (see \citet{Quiroz18,Nemeth21}), 
but this is in general model-specific.

At its core, the problem of working with large-scale data efficiently is a
question of how to exploit \emph{redundancy} in the data. To draw principled conclusions about a large data set based on a small
fraction of examples, one must rule out the presence of unique or interesting
additional information in the (vast) remainder of unexamined data.
One approach incorporates redundancy directly into its formulation: 
\emph{Bayesian coresets} \cite{Huggins16}.
The key idea is to represent the large-scale data by a small, weighted subset. The coreset can then be passed to any standard (automated) inference algorithm, 
providing posterior inference at a reduced computational cost.

Coresets come with a number of compelling advantages.  First, and
perhaps most importantly, \textbf{coresets preserve important model structure}.
If the original Bayesian posterior distribution exhibits symmetry, weak
identifiability, discrete variables, heavy tails, low-dimensional subspace
structure, or otherwise, the coreset posterior typically will exhibit that same
structure, because it is constructed using the same likelihood and prior as the
original model.  This makes coresets appealing for use in complex models where,
e.g., a Gaussian asymptotic assumption is inappropriate. Second,
\textbf{coresets are composable}: coresets for two data sets can often be
combined trivially to form a coreset for the union of data sets \cite{Feldman11}. This
makes coresets naturally applicable to streaming and distributed contexts \cite[Section 4.3]{Campbell19JMLR}.  Third, \textbf{coresets are
inference algorithm-agnostic}, in the sense that once a coreset is built, it
can be passed to most downstream inference methods---in
particular, exact MCMC methods with guarantees---with enhanced scalability.
Finally, \textbf{coresets tend to come with guarantees} relating the size of
the coreset to the quality of posterior approximation.

In this vignette, we will cover the basics of Bayesian coresets 
as well as recent advances in \cref{sec:introcoresets,sec:coresetadvances},
and discuss open problems and 
exciting directions for future work in \cref{sec:futurecoresets}.

\subsection{Introduction to Bayesian coresets}\label{sec:introcoresets}
\subsubsection{Setup}
We are given a target probability density $\pi(\theta)$ for 
$\theta\in \Theta$ that that is comprised of $N$ potentials
$\left(f_n(\theta)\right)_{n=1}^N$ and a base density $\pi_0(\theta)$,
\[
\pi(\theta) = \frac{1}{Z} \exp\left(\sum_{n=1}^N f_n(\theta)\right) \pi_0(\theta),\label{eq:coreset_base_model}
\]
where the normalization constant $Z$ is not known.
This setup corresponds to a Bayesian statistical model with prior $\pi_0$ and
i.i.d.~data $X_n$ conditioned on $\theta$, where $f_n(\theta) = \log p(X_n | \theta)$. 
The goal is to compute or approximate expectations under $\pi$; in the Bayesian scenario,
$\pi$ is the posterior distribution.

A key challenge arises in the large $N$ setting. Bayesian posterior computation algorithms tend to become intractable.
For example, MCMC typically has computational complexity $\Theta(NT)$ to obtain $T$
draws, since $\sum_n f_n(\theta)$ (and often its gradient) needs to be evaluated at each step. 
In order to reduce this $\Theta(NT)$ cost, \emph{Bayesian coresets} \cite{Huggins16}
replace the target with a surrogate density 
\[
\pi_w(\theta) = \frac{1}{Z(w)}\exp\left(\sum_{n=1}^N w_n f_n(\theta)\right) \pi_0(\theta),\label{eq:coreset}
\]
where $w\in\reals^N$, $w\geq 0$ are a set of weights,
and $Z(w)$ is the new normalizing constant. 
If $w$ has at most $M \ll N$ nonzeros, the $\Theta(M)$ cost of evaluating
$\sum_n w_n f_n(\theta)$ (and its gradient) is a significant improvement upon
the original $\Theta(N)$ cost.
The goal is then to develop an algorithm for coreset 
construction---i.e., selecting the weights $w$---that:
\benum
\item produces a small coreset with $M\ll N$, so that computation with $\pi_w$ is efficient;
\item produces a high-quality coreset with $\pi_w \approx \pi$, so that draws from $\pi_w$ are similar to those from $\pi$; and 
\item runs quickly, so that building the coreset is actually worth the effort for 
	subsequent fast draws from $\pi_w$.
\eenum
These three desiderata are in tension with one another. The smaller a coreset
is, the more ``compressed'' the data set becomes, and hence the worse the
approximation $\pi_w\approx \pi$ tends to be.  Similarly, the more efficient
the construction algorithm is, the less likely we are to find an optimal
balance of coreset size and quality with guarantees.

\subsubsection{Approaches to coreset construction}
There are three high-level strategies that have been used
in the literature to construct Bayesian coresets.

\paragraph*{\textbf{Subsampling}} The baseline method is to 
uniformly randomly pick a subset 
$\mcI \subseteq \{1,\dots,N\}$ of $|\mcI|=M$ data points and 
give each a weight of $N/M$, i.e.,
\[
w_n = \frac{N}{M} \quad \text{if } n \in \mcI, \quad w_n = 0 \text{ otherwise},
\]
resulting in the unbiased potential function approximation
\[
	\sum_{n=1}^N f_n(\theta) \approx N \left(\frac{1}{M} \sum_{m\in\mcI}f_m(\theta)\right).
\]
This method is simple and fast, but typically generates
poor posterior approximations. Constructing the subset by selecting
data with nonuniform probabilities does not improve
results significantly \cite{Huggins16}.
Empirical and theoretical results hint that in 
order to maintain a bounded approximation error, the subsampled coreset 
must grow in size proportional to $N$, 
making it a poor candidate for efficient large-scale inference.
Coresets therefore generally require more careful optimization.

\paragraph*{\textbf{Sparse regression}} 
One can formulate 
coreset construction as a sparse 
regression problem \cite{Campbell19JMLR,Campbell18,Zhang21b},
\[
\notag
w^\star = \argmin_{w\in\reals^N_+} \left\|\sum_{n=1}^N f_n - \sum_{n=1}^N w_nf_n\right\|^2 \quad\text{s.t.}\quad\|w\|_0\leq M,
\]
where $\|\cdot\|$ is some functional (semi)norm, 
and $\|w\|_0$ is the number of nonzero entries in $w$.
This optimization problem can be solved using iterative greedy optimization strategies
that provably, and empirically, provide a significant improvement in coreset quality over subsampling 
methods \cite{Campbell19JMLR,Campbell18,Zhang21b}.
However, this approach requires the user to design---and tends to be quite sensitive to---the
(semi)norm $\|\cdot \|$,  and so is not easy to use for the general practitioner.
The (semi)norm also typically cannot 
be evaluated exactly, resulting in the need
for Monte Carlo approximations with error that can dominate any improvement from
more careful optimization.

\paragraph*{\textbf{Variational inference}} Current state-of-the-art 
research  formulates the coreset construction problem as
variational inference in the family of coresets \cite{Campbell19},
\[
w^\star = \argmin_{w\in\reals_+^N} \mathrm{D}_{\mathrm{KL}}\left(\pi_{w} \| \pi\right)
\quad \text{s.t.}\quad\|w\|_0 \leq M.\label{eq:coresetklopt}
\]
Unlike the sparse regression formulation, this
optimization problem does not require expert user input.
However, it is not straightforward to evaluate the KL objective,
\[
&\log Z \!-\! \log Z(w) \!+\! \sum_{n=1}^N(w_n\!-\!1)\!\int \pi_w(\theta)f_n(\theta)\dee \theta,\label{eq:coresetkl}
\]
even up to a constant in $w$. The difficulty arises because \cref{eq:coresetkl} involves
 both the unknown normalization constant $Z(w)$ and an expectation
under $\pi_w$, from which we cannot in general obtain exact draws.
This is unlike a typical variational inference problem, 
where the normalization of the variational
density is known and obtaining draws is straightforward.
Current research on coreset construction is generally focused 
on addressing these issues; this is an active area of work, and 
a number of good solutions have been 
found \cite{Campbell19,Manousakas20,Jankowiak22,Naik22,Chen22,Manousakas22}.

\subsection{Notable recent advances} \label{sec:coresetadvances}
The literature on Bayesian coresets
is still in its early stages, and the field is developing quickly.
We highlight some key recent developments here.

\paragraph*{\textbf{Coreset data point selection}} Optimization-based coreset
construction methods have tended to take a ``one-at-a-time'' greedy selection
strategy to building a coreset, thus requiring a slow, difficult to tune
inner-outer-loop  \cite{Campbell19JMLR,Campbell19}.  Recent work
\cite{Chen22,Naik22,Jankowiak22} demonstrates coresets can be built 
without sacrificing quality by first uniformly subsampling the data set to select
coreset points, followed by batch optimization of the weights.
This is both significantly simpler and faster than past one-at-a-time selection approaches, while providing theoretical guarantees: for models with a strongly log-concave 
or exponential family likelihood, after subsampling, the KL divergence of the \emph{optimally-weighted} coreset 
posterior converges to 0 as $N\to\infty$ as long as the coreset size $M \gtrsim \log N$ \cite{Naik22}. 
This guarantee does not say anything about whether one can \emph{find} the optimal weights,
but just that selecting coreset data points by subsampling does not limit achievable quality. 

\paragraph*{\textbf{Optimizing the KL divergence}}

Given a selection of coreset points, there remains the problem of optimizing the KL
objective over the coreset weights $w$; this is challenging because one cannot 
obtain exact draws from $\pi_w$, or compute its normalization constant.
It is possible to use MCMC to draw from $\pi_w$,
and to circumvent the normalization constant issue by noting that derivatives are available
via moments of the potential functions under $\pi_w$, e.g.,
\[
\frac{\partial}{\partial w_n}\mathrm{D}_{\mathrm{KL}}\left(\pi_{w} \| \pi\right)
=&-\operatorname{Cov}_{w}\left[f_n(\theta), \sum_{i=1}^N (1-w_i)f_i(\theta)\right], \label{eq:klgrad}
\]
where $\operatorname{Cov}_{w}$ denotes covariance under $\pi_{w}$ \cite{Campbell19,Naik22}. 
The key difficulty of this approach is that it requires tuning the
MCMC method at each optimization iteration, as the weights $w$ (and hence the target $\pi_w$) 
are changing. Second order methods reduce the number of optimization iterations required
significantly \cite{Naik22}, and hence the challenge posed by needing to tune MCMC.

Another promising approach is to use a surrogate variational family
that is parametrized by the coreset weights $w$ but
 enables tractable draws and exact normalization constant evaluation \cite{Chen22,Jankowiak22,Manousakas22}.
For example, \citet{Chen22} propose using a variational surrogate family $q_w$ 
such that for all $w$, $q_w \approx \pi_w$, and then
optimizing the surrogate objective function
\[
	w^\star = \argmin_{w} \mathrm{D}_{\mathrm{KL}}\left(q_{w} \| \pi\right).
\]
\citet{Chen22} set $q_w$ to be a normalizing flow based on sparse Hamiltonian dynamics targeting $\pi_w$.
Concurrent work by \citet{Jankowiak22} proposes a similar idea, but based on 
variational annealed importance sampling \cite{Salimans15} as opposed to normalizing flows.
In either case, the optimization problem is then just a standard KL minimization over parameters $w$.
\citet{Manousakas22}, in contrast, propose using a generic variational family $q_\lambda$
parametrized by some auxiliary parameter $\lambda$ to take draws, and adds an additional penalty
to the optimization objective to tune $q_\lambda$ to approximate $\pi_w$:
\[
	w^\star, \lambda^\star = \argmin_{w,\lambda}  \mathrm{D}_{\mathrm{KL}}\left(\pi_w \| \pi\right) + \mathrm{D}_{\mathrm{KL}}\left(q_\lambda \| \pi_w\right).
\]
The unknown normalization constant on $\pi_w$ cancels in the two KL divergence terms,
and the $\mathrm{D}_{\mathrm{KL}}\left(\pi_w \| \pi\right)$ term is estimated using self-normalized importance sampling based
on draws from $q_\lambda$ (which should be close to $\pi_w$, ideally, due to the additional penalty term).
\citet{Manousakas22} use a diagonal-covariance Gaussian family for $q_\lambda$, and
use an inner-outer loop optimization method in which the inner loop optimizes $\lambda$
to help ensure that $q_\lambda$ remains close to $\pi_w$. 

These two approaches are strongly connected. Consider the optimal auxiliary 
parameter 
\[
	\lambda^\star(w) = \argmin_{\lambda} \mathrm{D}_{\mathrm{KL}}\left(q_\lambda \| \pi_w\right),
\]
and assume that the family $q_\lambda$ is flexible enough such that $q_{\lambda^\star(w)} = \pi_w$ for all $w$.
Then the two approaches are equivalent if we define $q_w = q_{\lambda^\star(w)}$:
\[
\mathrm{D}_{\mathrm{KL}}\left(\pi_w \| \pi\right) + \mathrm{D}_{\mathrm{KL}}\left(q_\lambda^\star(w) \| \pi_w\right)= \mathrm{D}_{\mathrm{KL}}\left(q_w \| \pi\right).
\]
The advantage of using a generic family $q_\lambda$ is that it is much easier
(and more flexible) than being forced to design a family $q_w$ satisfying
$q_w\approx \pi_w$. But self-normalized importance sampling is well-known to
often work poorly \cite{Chatterjee18} even when the reverse KL divergence is small,
and we still need to take draws from $\pi_w$ once the coreset is built.  The
approach of directly designing $q_w$ requires more up front effort, but the
optimization is well-behaved, and one can obtain \iid draws directly from $q_w$
afterward.

The tradeoff between the three current state-of-the-art approaches---second-order 
methods with draws from $\pi_w$ using MCMC \cite{Naik22},
direct surrogate variational methods with $q_w\approx \pi_w$ \cite{Chen22}, and
parametrized surrogate variational methods using 
$q_\lambda \approx \pi_w$ \cite{Manousakas22}---has not yet been explored empirically,
and is an open direction for future research.

\paragraph*{\textbf{Optimization guarantees}}
Although variational inference in general is nonconvex, the coreset variational
inference problem \cref{eq:coresetklopt} 
facilitates guarantees. In particular, 
\citet{Naik22} obtain geometric convergence to a point near the optimal coreset
via a quasi-Newton optimization scheme:
\[
\|w_k - w^\star_k\| &\leq \eta^k \|w_0 - w^\star_0\| + C,
\]
where $w_k$ is the $k^\text{th}$ iterate, and $w^\star_k$ is its 
projection onto a subset of optimal coreset weights (the optimum may not be unique).
The constants $\eta$ and $C$ are related to how good of 
an approximation the \emph{optimal} coreset is. 
If the optimal coreset is exact, then $0 < \eta < 1$ and $C = 0$.

\subsection{Open questions and future directions}\label{sec:futurecoresets}
Recent advances in coreset construction methods and theory have paved 
the way for a variety of new developments.
In this section we highlight important open problems and
areas for investigation.

\paragraph*{\textbf{Complex model structure, data, and symmetry}}
The coresets methodology and theory is now starting to coalesce for the basic
model setup in \cref{eq:coreset_base_model} with a finite-dimensional parameter
and conditionally \iid data.
Many popular models do not fit into this framework, such as 
certain network models \cite{Holland83},
continuous time Markov chains \cite{Anderson91}, etc.
Some of these models involve computational cost that scales poorly in $N$---e.g., Gaussian process 
regression with $O(N^3)$ complexity \cite{Williams95}---and would greatly
benefit from a summarization approach.
Even some models that technically fit in the framework of \cref{eq:coreset_base_model},
such as certain hierarchical models \cite{Blei03}, may be better
summarized if more of their latent structure is exposed to the coreset construction algorithm.

Moving beyond the conditionally \iid data setup, we advocate thinking about
this problem as \emph{model and data summarization}, broadly construed, as
opposed to just the specific case of coresets. At an abstract level, Bayesian
coresets are just one particular example of how one can construct a computationally inexpensive
parametrized variational family $\pi_w$ that provably contains (a distribution
near) the true posterior $\pi$. In general, there is no reason this has to be associated with
a sparse, weighted subset of data; we could, e.g., summarize networks with
subgraphs \cite{Orbanz17}, summarize high-dimensional data with low-dimensional
sketches \cite{Mahoney11}, summarize expensive, complicated neural network
structures with simpler ones \cite{ONeill20}, summarize expensive matrices with
low-rank randomized approximations \cite{Williams01}, etc.
The major question to answer is:

\emph{What is the natural extension of coresets, or
summarization more broadly, to more sophisticated models beyond
\cref{eq:coreset_base_model}? Is there a common underlying principle, or is
efficient summarization a problem that must be solved in a case-by-case manner?}

We believe that the key to answering these questions is to understand the connections
between Bayesian coresets, subsampling, probabilistic symmetries, and sufficiency in statistical
models; see, e.g., \cite{Diaconis88,Lauritzen88,Orbanz15}. Indeed, the fact that Bayesian coresets 
work at all is a reflection of the fact that one can use a small subset of data potentials as 
``approximately sufficient statistics,'' combined with the symmetry of their generating process.
Assuming a fruitful connection is made, we expect that current Bayesian 
coreset construction methods---which are based on 
subsampling to select a ``dictionary'' of potentials, followed by optimization to tune the 
approximation---will serve as a good template in more general models.

\paragraph*{\textbf{Improved surrogates and optimization}}
Early Bayesian coresets literature
\cite{Huggins16,Campbell18,Campbell19,Campbell19JMLR} suffered from the 
requirement of taking draws from $\pi_w$ both during and after construction. 
Sampling \emph{during} construction poses a particular challenge: if one intends to use
MCMC to take draws from $\pi_w$, one needs to continually adapt the MCMC kernel 
to a changing target $\pi_w$ as the weights $w$ are refined.
Recent developments discussed in \cref{sec:coresetadvances} suggest that an easier
way to approach the problem is to construct a tractable variational family
$q_w$ such that $q_w \approx \pi_w$ for all weights $w$---whether that is a normalizing
flow \cite{Chen22}, a variational annealed importance distribution \cite{Jankowiak22}, or 
an optimized parametric surrogate \cite{Manousakas22}---and then to tune the weights $w$
so that $q_w \approx \pi$. The benefit of this approach is the ability to take exact \iid draws 
and evaluate the density, which circumvents challenges of adaptive in-the-loop MCMC tuning.
This leads to the following question:

\emph{How should we construct a tractable, summarization-based variational family
such that $q_w \approx \pi_w$ for all $w$?}

For methods based on parametric surrogates \cite{Manousakas22} that set
$q_w = q_\lambda^\star(w)$, where $\lambda^\star(w) = \argmin_\lambda \mathrm{D}_{\mathrm{KL}}\left(q_\lambda \| \pi_w\right)$,
there are two major avenues for improvement. The first---and more likely achievable---goal
is in the optimization of the parametric surrogate. In particular, the methodology currently involves
slow inner-loop optimization of the surrogate, as well as potentially high-variance gradient
estimates based on self-normalized importance sampling. Handling these two issues would be 
a major step forward for this approach.
The second important area for future work---which may be far more challenging---is 
to provide theoretical guarantees on the quality of the coreset that is constructed using this method. 
The primary difficulty is that the surrogate optimization is as hard to analyze as other generic
variational inference problems.

For methods based on direct surrogates \cite{Jankowiak22,Chen22} where $q_w\approx \pi_w$ for all
$w$, there are again two major areas for improvement. First, current methods involve Hamiltonian
dynamics, and so are limited in scope to models with multidimensional real-valued variables; future
work should extend these methods to models with a wider class of latent variables.
The second area is once again to obtain rigorous theoretical guarantees on the quality of the surrogate.
This is likely to be much easier than in the general parametric surrogate case above, as $q_w$ is designed
to approximate $\pi_w$ directly, as opposed to just being a stationary point of a nonconvex optimization problem. 

\paragraph*{\textbf{Privacy, pseudo-data, and distributed learning}}
Distributed (or federated) learning is a task in which data are stored in separate data centers, and the goal
is to perform a global inference given all the data under the constraint that the data 
are not transmitted between centers. Both exact \cite{Dai19,Chan21} and
approximate \cite{Scott16,Broderick13} methods exist to perform Bayesian inference
in this setting. A common additional constraint is that the data within
each center are kept private, in some sense, from the other centers.

Coresets provide a potentially very simple solution to the 
distributed learning problem (both standard and privacy-preserving). 
In particular, coresets are often \emph{composable}: if one builds
subcoresets (independently and without communication) for subsets of a data set,
one can combine these trivially to obtain a coreset for the full data set \cite{Feldman20}.
Coresets have also been extended to the privacy-aware setting, where one
either trains pseudopoints with a differentially private scheme \cite{Manousakas20} 
or appropriately noises the coreset before sharing \cite{Feldman09}.
Subsequently, the data centers can share their privatized
summaries freely with one another, or with a centralized repository that
performs inference. There is some initial work on distributed Bayesian coresets constructed
via sparse regression techniques \cite[Section 4.3]{Campbell19JMLR},
but this work was done prior to the advent of modern construction methods.
Beyond this, there is no study in the literature
dedicated to theory and methods for distributed Bayesian coresets,
either privacy-preserving or otherwise.

\emph{How do we leverage recent advances in coreset construction to efficiently construct
differentially-private Bayesian coresets suitable for distributed learning
problems? What theoretical guarantees on communication cost and coreset
quality are possible?} 

\paragraph*{\textbf{Amortized and minimax coreset construction}}
Bayesian coresets are currently constructed in a model-specific manner
by minimizing the KL objective in \cref{eq:coresetklopt}.
In situations where multiple models are under consideration---in exploratory
analysis or sensitivity analysis, for example---one would need to re-tune 
the coreset weights for each model under consideration. Given that these
re-tuning problems all involve the same data, they should be closely related;
but it is currently an open question how to construct multiple related
coresets efficiently. In particular:

\emph{How generalizable are coresets? Is there a way to construct one optimized 
coreset that is appropriate for multiple models? Is there a way to 
amortize the cost of constructing multiple coresets for multiple models?}

One potential direction of future work is to formulate a minimax optimization
problem that is similar to \cref{eq:coresetklopt}, but where there is an outer maximization
over a set of candidate models. A major question along these lines is whether
it is actually possible to summarize a data set with a single coreset of $M \ll N$ data points such that 
the coreset provides a reasonable approximation for the worst-case model under consideration.
Another possible way to tackle the problem is to amortize the cost of multiple coreset construction,
in the spirit of \emph{inference compilation} \cite{Le17}.
Rather than constructing individual coresets, we train a ``coreset construction artifact:''
a function that takes as input a candidate model and data subsample, 
and outputs a set of coreset weights. In other words, we \emph{learn how to construct coresets}.
The most likely candidate for such an artifact is a recurrent deep neural network,
as is commonly-used in methods like inference compilation. 
A major question about this direction to consider is in which data analysis scenarios the cost 
of building such an artifact is worth the subsequent fast generation of coreset weights.

\paragraph*{\textbf{High-dimensional data and models}}
The coresets approach is designed with a focus on large-scale problems in the sense of 
the number of data points, $N$. But in practice, modern large-scale problems tend to also
involve high-dimensional data and latent model parameters; the dimension may even grow with $N$.
Empirical results have shown that coresets can be effective in problems 
with $10$-$100$-dimensional data and parameters,
while \emph{pseudo}coresets \cite{Manousakas20,Manousakas22}---which involves 
summarizing data with synthetic pseudodata points---have been 
used successfully on larger problems with $60{,}000$-dimensional parameters and $800$-dimensional data.
But results in this domain are limited, which leads to the following questions for future work:

\emph{When do we expect the coresets approach to work with high-dimensional data
and high-dimensional model parameters in general? Is there any modification to the (pseudo)coresets
approach required to achieve rigorous guarantees in this setting? How does the difficulty
of the coreset weight optimization scale in high dimensions?}

We begin with a negative (albeit pathological) example. When a large fraction of the 
potential functions $(f_n)_{n=1}^N$ encode unique
information in the posterior, the 
coresets approach breaks down; it is not possible to maintain a good posterior approximation upon
removing potentials. \citet[Proposition 1]{Manousakas20} makes this intuition
precise with a simple example. In a
$d$-dimensional Gaussian location model with prior $\theta \distas \distNorm(0,I)$, 
likelihood $\distNorm(\theta,I)$, and data generated via
$X_n\distiid \distNorm(0,I)$, the \emph{optimal} coreset of any size $M < d$
satisfies 
\[
	\mathrm{D}_{\mathrm{KL}}\left(\pi_{w^\star} \| \pi\right) \gtrsim d\quad\text{as}\quad d\to\infty,
\]
with high probability\footnote{The result by \citet{Manousakas20} is stated in terms of the inverse CDF
of a $\chi^2$ distribution with $d-M$ degrees of freedom. The $\Omega(d)$ lower bound follows directly
by noting that $X\distas\distChiSq(d-M)$ implies 
\[
	\frac{X-(d-M)}{\sqrt{2(d-M)}} \convd \distNorm(0,1) \qquad d\to\infty.
\]
}. In some sense, this is unsurprising; the Gaussian location model 
with large $d$, despite is mathematical simplicity, is a worst-case scenario for data
summarization, as one needs at least $d$ potential functions $f_n$ to span a $d$-dimensional space.

But in practice, high-dimensional data do not typically exhibit this worst-case behaviour;
they often instead exhibit some simpler, lower-dimensional structure.
Developing (pseudo)coreset methods that take advantage of that structure is a key step needed
to make summarization a worthwhile approach in large-scale modern problems.
Furthermore, assuming that the coreset size should generally increase with dimension,
additional work is needed to understand how the difficulty of the stochastic weight optimization
scales. It is worth investigating whether the recently-developed literature on data
distillation in deep learning \cite{Wang18b} contains any insights applicable to the
Bayesian setting.

\paragraph*{\textbf{Improved automation and accessibility}} Recent advances in
research have, for the first time, made coresets a practical approach
to efficient Bayesian computation. However, there is still much work to do to
make their use possible by nonexperts. First and foremost,
there is a need to develop a general, well-engineered code base that
interfaces with common probabilistic programming libraries like Stan 
and Turing \cite{carpenter2017stan,Ge18}.
In addition, there is a need for
automated methods to (a) select coreset weight optimization tuning parameters,
(b) select coreset size, and (c) assess and summarize the quality of the
coreset.

\paragraph*{\textbf{Other divergences}}
Currently, all variational coreset construction approaches optimize
the reverse Kullback-Leibler divergence. A straightforward direction for 
future work would be to investigate the effect of using alternative
divergences, e.g.~the R\'enyi divergence \cite{Li16} or $\chi^2$ divergence \cite{Dieng17},
in \cref{eq:coresetklopt}. These will all likely pose similar issues 
with the unknown normalization constant $Z(w)$, but divergences other
than the reverse KL may provide coresets with distinct statistical properties.

\section{Distributed Bayesian Inference}

Distributed methods for Bayesian inference address the challenges posed by massive data using a divide-and-conquer technique. They exploit distributed computing to reduce the time complexity of Monte Carlo algorithms that require multiple sweeps through the data in every iteration. During the last decade, three main groups of distributed methods have been developed for Bayesian inference. The first class of methods is the simplest and has three steps: divide the data into disjoint subsets and store them across multiple machines, run a Monte Carlo algorithm in parallel on all the machines, and combine parameter draws from all the subsets on a central machine. The last step requires one round of communication, so these approaches belong to the class of \emph{one-shot learning} methods \citep{WanDun13,Minetal14,Neietal14,Srietal15,Wanetal15,Scoetal16,Minetal17,NemShe18,SavSri18,GuhBan18,Dai19,XueLia19,Joretal19,WuRob17,Min19,Guhetal17,Guhetal19,Mesetal20,Deetal22}. They are based on a key insight that the subset parameter draws provide a noisy approximation of the true posterior distribution and differ mainly in their combination schemes.  

The second class of methods relies on distributed extensions of stochastic gradient MCMC \citep{Ahnetal14, Kor14,Cheetal16, Eletal21}, which are typically based on stochastic gradient Langevin dynamics (SGLD) \citep{Welling11,Maetal15}. They also split the data into subsets but have several rounds of communication among the machines. In every iteration, they select a subset with a certain probability, draw the parameter using a modified SGLD update, and communicate the parameter draw to the central machine. The high variance of the stochastic gradients and high communication costs have motivated the development of the third set of methods \citep{Broetal18,Plaetal21}.
They are stochastic extensions of global consensus methods for distributed optimization, such as Alternating Direction Method of Multipliers (ADMM) \citep{Boy11, ParBoy14}. They divide the data into subsets, store them on machines, and augment the posterior density with auxiliary variables. These variables are conditionally independent given the parameter, and the parameter's marginal distribution reduces to the target under certain limiting assumptions. The former assumption is crucial for drawing the auxiliary variables in parallel, whereas the latter condition ensures asymptotic accuracy. Every iteration consists of synchronous updates where the machines storing the data draw the auxiliary variables and send them to the central machine that uses them to draw the parameter \citep{Von19, Renetal20, Von20, Plaetal21,Von22}.

Distributed Bayesian methods have three main advantages. First, most of them are algorithm-agnostic and are easily used with any Monte Carlo algorithm. 
Second, distributed methods come with asymptotic guarantees about their accuracy. Such results show that approximated and target posterior distributions are asymptotically equivalent under mild regularity assumptions. Finally, they are easily extended to handle application-specific constraints, such as clustering of samples in nonparametric models \cite{Ni20} and privacy-preserving federated learning \citep{kidd2022federated}.

We cover the basics of distributed Bayesian inference and recent advances in \cref{sec:one-shot}--\cref{sec:axda},  
and discuss future research directions in \cref{sec:future-dnc}.

\subsection{One-shot learning}\label{sec:one-shot}

We provide a brief overview of one-shot learning
approaches for distributed Bayesian inference. There is a rich variety of such algorithms available in the literature. We start with the most common setup that assumes the  observations are conditionally independent given the parameter, leading to a product form for the likelihood. Let $Y_1^N = (Y_1, \ldots, Y_N)$ denote the observed data. The model is specified using  the distribution $\PP_{\theta}$ with density $p(y \mid \theta)$ and $p$-dimensional parameter $\theta \in \Theta \subseteq \RR^p$. Assume that  $Y_1^N$ are randomly partitioned into $K$ disjoint subsets. Let $Y_{(j)} = \{Y_{(j)1}, \ldots, Y_{(j)M}\}$ be the $j$th subset ($j=1, \ldots, K$), where we have assumed that all the subset sample sizes equal $M$ for simplicity. The true and subset $j$ likelihoods are $\ell_N(\theta) = \prod_{i=1}^N p(Y_i \mid \theta)$ and $\ell_{jM}(\theta) = \prod_{i=1}^M p(Y_{(j)i} \mid \theta)$. Let  $\Pi$ be a prior distribution on $\Theta$ with density $\pi(\theta)$. Then, the posterior density of $\theta$ given $Y_1^N$ is $\pi_N(\theta \mid Y_1^N) = \ell_{N}(\theta) \pi (\theta) / C_N$, where $C_N=\int_{\Theta} \ell_{N}(\theta) \pi(\theta) d \theta$ and $C_N$ is finite.

\paragraph*{\textbf{Consensus Monte Carlo (CMC) and its generalizations}} These methods exploit the observation that the full data posterior can be factored as a product of subset posteriors with tempered priors \citep{Scoetal16}:
\begin{align}\label{eq:cmc}
    \pi_N(\theta \mid Y_1^N) &=  C_N^{-1} \prod_{j=1}^K \{\pi (\theta)\}^{1/K} \ell_{jM}(\theta) \nonumber \\
    &\propto \prod_{j=1}^K \pi_M(\theta \mid Y_{(j)}) \equiv \prod_{j=1}^K \pi_{j}(\theta).
\end{align}
Here $\pi_M(\theta \mid Y_{(j)})$ (or $\pi_{j}(\theta)$) is the $j$th subset posterior density of $\theta$ computed using likelihood and prior $\ell_{jM}(\theta)$ and $\{\pi (\theta)\}^{1/K} $. Let $\theta_{(j)t}$ be the parameter draws obtained from $\pi_{j}(\theta)$ using a Monte Carlo algorithm ($j=1, \ldots, K$; $t = 1, \ldots, T$) and $\hat \pi_{j}(\theta)$ be an estimate of $\pi_{j}(\theta)$ obtained using $\theta_{(j)t}$s. Then, $\prod_{j=1}^K \hat \pi_{j}(\theta)$ is proportional to an estimate of  $\pi_N(\theta \mid Y_1^N)$.
In the special case that $\pi_{j}(\theta)$s are Gaussian, then so is $\pi_N(\theta \mid Y_1^N)$ and weighted average of $\theta_{(j)t}$s correspond to draws from $\pi_N(\theta \mid Y_1^N)$  \citep{Scoetal16}. More accurate combination algorithms estimate $ \pi_{j}(\theta)$ using kernel density estimation \citep{Neietal14}, Weierstrass transform \citep{WanDun13}, random partition trees \cite{Wanetal15}, Gaussian process regression \citep{NemShe18}, and normalizing flows \citep{Mesetal20}, where the last two approaches also use importance sampling to select promising $\theta_{(j)t}$s for better approximation accuracy.

\paragraph*{\textbf{Median and mean posterior distributions}} These methods combine the subset posterior distributions using their geometric center, such as
the median and mean posterior distributions. The main difference between them and CMC-type approaches is the definition of subset posterior densities. Specifically, the $j$th subset posterior density is 
\begin{align}\label{eq:dist}
    \pi_M(\theta \mid Y_{(j)}) =  C_M^{-1} \{ \ell_{j M}(\theta) \}^{K} \pi(\theta) \equiv \tilde \pi_{j}(\theta),
\end{align}
where $C_M = \int_{\Theta}  \{ \ell_{jM}(\theta) \}^{K} \pi(\theta) d \theta$ is assumed to be finite for posterior propriety. The pseudo-likelihood $\{ \ell_{jM}(\theta) \}^{K}$ in \eqref{eq:dist} is the likelihood of a pseudo sample resulting from replicating every sample in the $j$th subset $K$ times \citep{Minetal14}. This pseudo-likelihood ensures the posterior variance of the subset and true posterior densities are calibrated up to  $o_{P}(N^{-1})$ terms \citep{Lietal17,Minetal17,Srietal18}. Similar to the previous methods, $\theta_{(j)t}$s are drawn in parallel from $\tilde \pi_{j}(\theta)$s using any Monte Carlo algorithm. Let $\tilde \Pi_j$ be the $j$th subset posterior distribution with density $\tilde \pi_{j}(\theta)$. Then, its empirical approximation supported on the $\theta_{(j)t}$s is $\hat \Pi_j = T^{-1}\sum_{t=1}^T \delta_{\theta_{(j)t}}(\cdot)$, where $\delta_{\theta}(\cdot)$ is the delta measure supported on $\theta$. The median and mean  posterior distributions are approximated using empirical measures $\hat \Pi^*$ and $\hat {\overline \Pi}$ that are supported on $\theta_{(j)t}$s. The weights of $\theta_{(j)t}$s are estimated via optimization such that $\sum_{j=1}^K \mathsf{d}(\hat \Pi^*, \hat \Pi_j)$ and $\sum_{j=1}^K \mathsf{d}^2(\hat {\overline \Pi}, \hat \Pi_j)$ are minimized, respectively, where $\mathsf{d}$ is a metric on probability measures \citep{Minetal14,Srietal15}. If $\theta$ is one dimensional and $\mathsf{d}$ is the $2$-Wasserstein distance, then the $\alpha$th quantile of the mean posterior  equals the average of $\alpha$th quantiles of the $K$ subset posteriors
\citep{Lietal17}.


\paragraph*{\textbf{Mixture of recentered subset posteriors}} 
The final combination algorithm uses a $K$-component mixture of re-centered subset posterior densities  in \eqref{eq:dist}. Let $\overline \theta_{(j)}$ be the mean of 
$\pi_M(\theta \mid Y_{(j)})$ and $\overline \theta = \sum_{j=1}^K \overline \theta_{(j)} / K$. Then, the distributed posterior distribution with density
\begin{align}
  \label{eq:3}
  \tilde \pi(\theta \mid Y_1^N)  = \sum_{j=1}^K \frac{1}{K} \tilde \pi_{j}(\theta - \overline \theta + \overline \theta_j),
\end{align}
approximates $\pi_N(\theta \mid Y_1^N)$, where  $\tilde \pi_{j}$ is defined in  \eqref{eq:dist} \citep{XueLia19,WuRob17}. To generate draws from $\tilde \pi(\theta \mid Y_1^N)$ in \eqref{eq:3}, we obtain the  empirical approximation of the distributed posterior $\tilde \Pi$ with density $\tilde \pi(\theta \mid Y_1^N)$ as
\begin{align}
  \label{eq:5}
  \hat{\tilde \Pi} = \sum_{j=1}^K \sum_{l=1}^T \frac{1}{KT} \delta_{\hat \theta + \theta_{(j)l} - \hat \theta_j}(\cdot),    
\end{align}
where  $\hat \theta_j = \sum_{l=1}^T \theta_{(j)l} / T$ and $\hat \theta = \sum_{j=1}^K \hat \theta_j / K$.   
The $K$-mixture $\hat{\tilde \Pi} $ and geometric centers $\hat \Pi^*, \hat {\overline \Pi}$  are similar in that the atoms of the empirical measures are transformations of the subset posterior draws. The main difference between them lies in their approach to estimating the weights of the atoms. All the atoms of $\hat{\tilde \Pi} $ have equal weights (i.e., $(KT)^{-1}$), whereas the atom weights of $\hat \Pi^*$  and $\hat {\overline \Pi}$ are non-uniform and estimated via an optimization algorithm.

\paragraph*{\textbf{Asymptotics}} The large sample properties of the posterior estimated in one-shot learning, denoted as $\Pi_{\text{D}, N}$, are justified via a Bernstein-von Mises (BvM) theorem; however, these results are only known for the last two methods and not for the CMC-type approaches \citep{Lietal17, Minetal17}.  A BvM for  
$\Pi_{\text{D}, N}$ shows that it is asymptotically normal under mild assumptions as $K$ and $N$ tend to infinity. The center of the limiting distribution is specific to the combination algorithm, but the asymptotic covariance matrix equals $I_0/N$, where $I_0$ is the Fisher information matrix computed using $Y \sim \PP_{\theta_0}$. This shows that the asymptotic covariance of the true and distributed posteriors are calibrated up to $o_P(N^{-1})$ terms. Under these assumptions, 
\begin{align}
  \label{eq:8}
 \left\| \Pi_{\text{D}, N} (\cdot \mid Y_1^N) - \Pi_N (\cdot \mid Y_1^N)  \right\|_{\text{TV}} \leq \|\tilde \theta - \hat \theta  \|_{2} 
\end{align}
ignoring $ o_P(N^{-1/2})$ terms, where $\| \cdot \|_{\text{TV}}$ is the total variation distance, $\hat \theta$ is the maximum likelihood estimate (MLE) of $\theta$ computed using $Y_1^N$, and $\tilde \theta$ is a center of the $K$ subset MLEs  of $\theta$: $\hat \theta_1, \ldots, \hat \theta_K$. They satisfy
$\|\hat \theta_j - \theta_0 \|_2 = o_P(M^{-1/2})$, so 
$\|\tilde \theta - \theta_0 \|_2 = o_P(M^{-1/2})$ because $\tilde \theta$ is a center of the subset MLEs. Furthermore, $\|\hat \theta - \theta_0 \|_2 = o_P(N^{-1/2})$ and combining it with the previous result implies that $\|\tilde \theta - \hat \theta  \|_{2} = o_P(M^{-1/2})$, which does not scale in $K$. This shows that the bias of $\Pi_{\text{D}, N}$ in approximating $\Pi_N $ does not decrease as $K$ increases, and  that $K$ does not generally impact the approximation accuracy of $\Pi_{\text{D}, N}$, unless $\tilde \theta$ is a root-$N$ consistent estimator of $\theta_0$.

\paragraph*{\textbf{Notable recent advances}}
One-shot learning, except CMC-type methods, has been generalized to dependent data. 
In time series data, smaller blocks of consecutive observations form the subsets and preserve the ordering of samples. A measure of dependence, such as the mixing coefficient, dictates the choice of $K$. The subset pseudo-likelihood in \eqref{eq:dist} is modified to condition on the immediately preceding time block to model the dependence and raised to a power of $K$. For one-shot learning in hidden Markov models with mixing coefficient $\rho$, the distributed posterior estimated using  \eqref{eq:5} with the modified pseudo-likelihood and $K = o(\rho^{-M})$ satisfies \eqref{eq:8} \citep{WanSri21}. These results have been generalized to a broader class of  models, but guidance on the choice of $K$ remains underexplored \citep{Ouetal21}. 

Posterior computations in Gaussian process (GP) regression fail to scale even for moderately large $N$ \citep{RasWil06,Banetal14}. One-shot learning
has addressed this challenge but with no theoretical results \citep{Minetal14,Srietal15}. The choice of $K$ here depends on the smoothness of the regression function. Assuming a higher order of smoothness of regression functions guarantees accurate estimation on the subsets for larger values of $K$. Specifically, if the regression function is infinitely smooth, the predictor lies in $[0,1]$, and $K =  O(N / \log^2 N)$, then the decay rates of estimation risks for the distributed and true posterior distributions depend only on $N$ and are asymptotically equivalent. In more general problems where the regression function belongs to the H\"older class of functions on $[0, 1]^D$ with smoothness index $\alpha$, the upper bound for $K$ depends on $N, D,$ and $\alpha$ for guaranteeing optimal decay rate of the estimation risk  \citep{Guhetal17}. These results have been generalized to varying coefficients models \citep{Guhetal19}.

\paragraph*{\textbf{Limitations}}
The main limitation of one-shot learning methods is their reliance on the normality of the subset posterior distributions. Scaling of the parameter draws on the subsets helps in some cases but fails to generalize beyond the family of elliptical posterior distributions \citep{ShySri22,Vynetal23}. \citep{Deetal22} identify three additional problems for one-shot learning. First, subset posteriors fail to capture the support of a multimodal posterior with a high probability.  Second, a subset posterior can be substantially biased and fail to be a reasonable approximation of the true posterior, violating another major assumption. Finally, subset posterior draws fail to provide information about the tails of the true posterior, resulting in poor estimates of tail event probabilities.  A key observation of \citep{Deetal22} is that communication among machines is necessary for improving the approximation accuracy of subset posteriors.

\subsection{Distributed stochastic gradient MCMC}\label{sec:axda}

Langevin Monte Carlo uses the gradient of $\log \pi_N( \theta \mid Y_1^N)$ for generating $\theta$ proposals in a Metropolis-Hastings sampling scheme \citep{neal2011mcmc}. The gradient computation requires cycling through all the samples, which is prohibitively slow for a large $N$. SGLD bypasses this problem by subsampling a size $n$ subset $S_n$ of $\{1, \ldots, N\}$ and proposing $\theta$ in the $(t+1)$th iteration given $\theta_t$ using a noisy approximation of the gradient $g_N(\theta) = \nabla \log \pi_N( \theta \mid Y_1^N)$ as follows: 
\begin{align}\label{grad}
\theta_{t+1} &= \theta_{t} + \frac{h_t}{2} \, \hat  g_n(\theta_t) + \epsilon_t, \quad \epsilon_t \sim \Ncal(0, h_t I), \\
     \hat  g_n(\theta) &= \nabla \log \pi(\theta) + \frac{N}{n}
    \sum_{i \in S_n} \nabla \log p(Y_i \mid \theta), \nonumber
\end{align}
where $\hat  g_n(\theta)$ is a noisy estimate of $g_N(\theta)$ in that 
\begin{align*}
    \EE [\hat  g_N(\theta) / N] \approx \EE [g_N(\theta) / N ]
\end{align*}
for every $\theta$. The step size  $h_t$ decreases to 0 such that $\sum_{t=1}^\infty h_t = \infty$ and $\sum_{t=1}^\infty h_t^2 < \infty$. The discretization error of the Langevin dynamics is negligible as $h_t \rightarrow 0$, so the rejection probability of $\theta_t$ in the Metropolis-Hastings step approaches 0 \citep{Welling11}. In practice, however, 
$h_t \propto 1/N$ for better mixing and efficiency \citep{Broetal18}. This produces a chain $\{\theta_t\}$ that does not have the target as the stationary distribution, but it mimics the true continuous-time Langevin dynamics closely and hence has ``approximately'' the right target.

The distributed SGLD extension (DSGLD) performs the SGLD update on randomly selected subsets \citep{Ahnetal14}. Let $p = (p_1, \ldots, p_K)$ be a vector of positive probabilities such that $p_1 + \ldots + p_K = 1$, and $p_j$ is the probability of selecting subset $j$ for the SGLD update. At the $(t+1)$th iteration, simple distributed SGLD extension selects a subset $j_t \sim \text{Categorical}(p)$ and defines
\begin{align}\label{dgrad}
    \theta_{t+1} &= \theta_{t} + \frac{h_t}{2} \, \hat  g_{mj_t}(\theta_t) + \epsilon_t, \quad \epsilon_t \sim \Ncal(0, h_t I),\\
    \hat  g_{mj_t}(\theta) &= \nabla \log \pi(\theta) + \frac{M}{p_{j_t} m}
    \sum_{i \in S_m} \nabla \log p(Y_{(j_t)i} \mid \theta),\nonumber
\end{align}
where  $S_m$ is a size $m$ subsample of $\{1, \ldots, M\}$. The chain  $\{ \theta_t \}$ jumps to the next worker selected for the SGLD update, and this process continues until convergence. This scheme is undesirable due to communication overload; therefore, DGLD samples $\theta$ using \eqref{dgrad} multiple times on the  selected subset before the chain $\{ \theta_t \}$ jumps to a new subset. Additionally, the communication bottlenecks are minimized by selecting ``optimal'' workers with minimum wait times before the chain $\{ \theta_t \}$ jumps; see Section 3.2 in \citep{Ahnetal14}.

The efficiency gains in DSGLD come at the cost of loss in asymptotic accuracy. The main reason is that the smaller subset sizes imply that the possible subsample combinations on a subset are much smaller than those obtained using the full data in the standard SGLD update. Better gradient surrogates with smaller variance and higher asymptotic accuracy have been developed \citep{Cheetal16, Eletal21}, but the variance of stochastic gradients increases with $N, K$, and data heterogeneity, resulting in convergence failures \citep{Broetal18}. 

\subsection{Asymptotically exact data augmentation}\label{sec:axda}

Asymptotically exact data augmentation (AXDA) generalizes DA using stochastic extensions of global consensus optimization algorithms such as ADMM \citep{Renetal20,Von19,Von20}. AXDA has 
subset-specific auxiliary variables $z = (z_1, \ldots, z_K) \in \prod_{k=1}^K \RR^{M}$ and tolerance parameter $\rho  \in \RR_{+}$, which are similar to ``missing data'' in DA and tolerance parameter in ADMM. Using the notation in \eqref{eq:cmc}, $z$ is chosen such that the augmented density satisfies
\begin{align}\label{eq:axda}
    \pi_{\rho}(\theta, z_1, \ldots, z_K \mid Y_1^N) \propto \pi(\theta) \prod_{k=1}^K \ell_{k,\rho}(\theta, z_k),
\end{align}
where $\ell_{k,\rho}(\theta, z_k) = p_k(z_k, Y_{(k)}) \kappa_\rho(z_k, \theta)$, $\kappa_\rho$ is a kernel such that $\kappa_\rho(\cdot, \theta)$ converges weakly to $\delta_{\theta}(\cdot)$ as $\rho \rightarrow 0$, and $p_k(z_k, Y_{(k)})$ is such that  $\lim_{\rho \rightarrow 0} \int \ell_{k,\rho}(\theta, z_k) \, dz_k = \ell_{kM}(\theta) = \prod_{i=1}^M p(Y_{(k)i}\mid \theta) $; that is, $z$ plays the role of missing data  and preserves the observed data model as $\rho \rightarrow 0$, justifying that  AXDA is  asymptotically exact.

The advantage of the density in \eqref{eq:axda} is that the $z_k$s are conditionally independent given $\theta$. In every iteration, $z_k$s are drawn in parallel on the machines storing $Y_{(k)}$s. These draws are communicated to the central machine that uses them to draw $\theta$ and generates a Markov chain for $\theta$, whose stationary density equals  $\pi_N(\theta \mid Y_1^n)$ under mild assumptions. AXDA has been used for Bayesian inference in generalized linear models and nonparametric regression \citep{Renetal20,Von20}, but proper choices of $p_k(z_k, Y_{(k)})$, $\rho$, and  $\kappa_\rho$ limit the broader application of AXDA. 
\citep{Von22} and \citep{Plaetal21} develop AXDA using ADMM-type variable splitting and  
Langevin Monte Carlo algorithms. Like DSGLD, repeated communications among the machines diminish the computation gains from distributed computations. 

\subsection{Open questions and future directions}\label{sec:future-dnc}
This section highlights the limitations of distributed inference methodology, important open problems, and areas for future investigation.

\paragraph*{\textbf{High dimensional and dependent data models}}
A variety of options exist for distributed Bayesian inference in independent data models, but they fail to generalize to high-dimensional models. The literature on distributed methods for inference in high dimensional models is sparsely populated \citep{Joretal19}. The development of distributed methods that exploit the low dimensional structure in high dimensional problems is desired. 

Most distributed methods assume that the likelihood has a product form; see \eqref{eq:cmc}. This assumption fails for many time series and spatial models.  There are one-shot learning methods for hidden Markov models \citep{WanSri21}, but they are inapplicable beyond the family of elliptical posterior distributions. No dependent data extensions are available for DSGLD and AXDA algorithms.  

\paragraph*{\textbf{Bias and variance reduction}} The bias between the true and distributed posterior in one-shot learning fails to decay as $K$ increases. For parametric models, (\ref{eq:8}) shows that the distributed distribution has a bias of the order $o_P(M^{-1/2})$, which is suboptimal compared to $o_P(N^{-1/2})$  order bias  of the true posterior. This means that increasing $K$  has no impact on the accuracy of the distributed posterior. One way to bypass this problem is by centering the distributed posterior at a root-$N$ consistent estimator; see \citep{WanSri21}. 
Addressing this problem  is useful for Bayesian federated learning, where one-shot learning is increasingly used due to its  simplicity \citep{kidd2022federated}. Similarly, developing gradient surrogates with smaller variances is crucial for Bayesian federated learning using Langevin Monte Carlo.

\paragraph*{\textbf{Asynchronous updates}} Synchronous updates are crucial for convergence guarantees of DSGLD and Langevin Monte Carlo algorithms based on AXDA; however, synchronous updates become expensive as the number of subsets increases, resulting in diminishing benefits of distributed computations. Asynchronous updates bypass such problems when the subset sizes are similar, but they imply that the $\{\theta_t\}$ chain is not Markov, which rules out conventional tools for proving convergence guarantees. Asynchronous DSGLD and AXDA extensions have numerous practical benefits. \citep{Zhouetal22} have developed asynchronous DA for variable selection and mixed effects model, but its extension to a broader class of models remains unknown.

\paragraph*{\textbf{Generalized likelihoods}}
Bayesian inference using generalized likelihoods has several advantages, including robustness and targeted inference; however, the current literature relies heavily on exploiting the structure of the hierarchical model. Preliminary results are available about the commonalities between AXDA and approximate Bayesian inference \citep{Von20}. For broader applications, it is interesting to explore distributed extensions of the \emph{cut}  posterior in  misspecified models  \citep{Plum15} and distributed inference in Bayesian models based on generalized likelihoods.

\paragraph*{\textbf{Applications}}
Distributed Bayesian inference has found applications in federated learning \citep{kidd2022federated}. These methods are ideal for Bayesian analysis of multi-center longitudinal clinical studies because the data cannot be moved to a central location due to  privacy concerns. Limited examples of such applications are available; therefore, it is interesting to explore such privacy-preserving extensions of distributed methods.

\paragraph*{\textbf{Automated diagnostics and accessibility}}
Automated application and model diagnostics for distributed methods have received little attention. One-shot learning methods are easily implemented using the parallel R package \citep{Par21}; however, a similar general purpose software for deploying the distributed algorithms in practice remains to be developed. Addressing these challenges is crucial in facilitating the wide applicability of distributed methods.

\section{Variational Bayes}

Although variational approximations are mentioned in passing within previous sections, in this section we provide a vignette focused specifically on Variational Bayesian (VB) methods, 
which approximate the posterior distribution by a member in a simpler class of distributions through minimizing the KL divergence.  Below, we review some recent developments on theory and computation for variational Bayes and outline future directions.  
 
 \subsection{Introduction to variational Bayes} 
 
We first describe our setup for a \textit{statistical experiment}, defined as a pair of a sample space and a set of distributions on the sample space. For each sample size $n\in\mathbb N$,  suppose that we observe a $\cX_n$-valued sample $\bX^{(n)}$, where $\cX_n$ is a measurable \textit{sample space} equipped with a reference $\sigma$-finite measure $\mu_n$.  The sample is modeled with a distribution $P_{\theta}^{(n)}\in\mathcal P(\bX_n)$ determined by a parameter $\theta$ in a measurable parameter space $\Theta_n$.  
 
 Let $\Pi(\theta)$ be a prior distribution  of $\theta$ on $\Theta_n$ which often comes with a prior density $\pi(\theta)$.   If a collection of distributions $\{\P_\theta:\theta\in\Theta_n\}$ is dominated by some measure $\mu,$ then Bayes's rule gives the posterior distribution
            \begin{equation*}
               \Pi( \d\theta|\bX_n)\propto \underbrace{\frac{\d \P_\theta}{\d\mu_n}(\bX_n)}_{\text{likelihood}}
                \underbrace{\Pi(\d\theta)}_{\text{prior}}.
            \end{equation*}
 
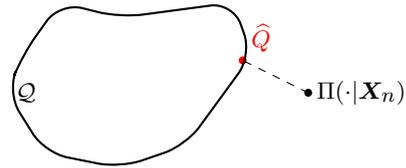
\begin{figure}
           \begin{tikzpicture}[>=latex]
                \coordinate (a) at (2.5,0);
   	             \coordinate (b) at (1.63,0.43);
                \fill (a) circle (1.5pt)  node[right] {$\Pi(\cdot|\bX_n)$};
                \fill[red] (b) circle (1.5pt)  node[above right] {$\hQ$};
                \draw[dashed] (a) -- (b);
                \draw[thick,rounded corners=3mm] (-1.5,0)--(-1.3,.5)--(-.8,1.2)--(0,1)--(1.5,1.2)--(1.75,.5)--(0.875,-.7)--(0,-1)--(-1,-.8)--cycle;
                \draw (-1.5,0) node[right] {$\cQ$};
            \end{tikzpicture}
            \caption{An illustration of variational Bayes}
             \label{fig-vb}
\end{figure}

Variational Bayes (VB) aims to provide an approximation to the posterior distribution $\Pi(\cdot | \bX_n)$. More specifically,  VB turns Bayesian computation into a trackable  optimization problem. To do this,  one first posits a family of approximate distributions  $ \cQ$ called a \emph{variational family}, which is a set of distributions on $\Theta$.
          The goal is then to find a member of the variational family that minimizes the KL divergence to the exact posterior $\Pi(\cdot|\bX_n)$:  
                \begin{equation}
                \label{eq-objective}
                \hQ=\argmin_{Q \in \cQ} \mathrm{D}_{\mathrm{KL}}\left(Q \| \Pi(\cdot|\bX_n)\right)
                \end{equation}
See Figure \ref{fig-vb} for a simple graphical illustration. 
The posterior is approximated with the optimal member $\hQ$ of the family, which is  called a \emph{variational posterior}. Statistical inference is then based on the variational posterior $\hQ$. 

        In solving the optimization problem \eqref{eq-objective}, one first writes $\Pi(\d\theta|\bX_n)=p_\theta(\bX_n)\Pi(\d\theta)/p(\bX_n)$, where $p(\bX_n):=\int p_\theta(\bX_n)\Pi(\d\theta)$ is the marginal likelihood of $\bX_n$.  The KL-divergence above can be written as
\[
&\mathrm{D}_{\mathrm{KL}}\left(Q \| \Pi(\cdot|\bX_n)\right)
 =\int \log \left(\frac{p(\bX_n)Q(\d\theta)}{p_\theta(\bX_n) \Pi(\d\theta)}\right) Q(\d\theta)\nonumber\\
&=\underbrace{-\int\log p_\theta(\bX_n) Q(\d\theta) + \mathrm{D}_{\mathrm{KL}}\left(Q \| \Pi\right)}_{=:\Psi(Q, \Pi, \bX_n)\; :=-ELBO} + \log p(\bX_n). \label{eq-elbo}
\]
In the above, we let  
$$\Psi(Q, \Pi, \bX_n)=-\int\log p_\theta(\bX_n) Q(\d\theta) + \mathrm{D}_{\mathrm{KL}}\left(Q \| \Pi\right),$$ 
which we call the \emph{variational objective function.}  This is also the negative of the \emph{evidence lower bound (ELBO)}, where the ELBO is  $\int\log p_\theta(\bX_n) Q(\d\theta)- \mathrm{D}_{\mathrm{KL}}\left(Q \| \Pi\right)$ which provides a lower bound of the `evidence' or the marginal likelihood $\log p(\bX_n)$ as seen from \eqref{eq-elbo}.
            
             Since $p(\bX_n)$ is a constant with respect to $Q$, one has 
            \begin{equation}
                \label{eq-vbopt}
                \hQ=\argmin_{Q\in \cQ}\Psi(Q, \Pi, \bX_n)=\argmin_{Q\in \cQ}\mathrm{D}_{\mathrm{KL}}\left(Q \| \Pi(\cdot|\bX_n)\right).
            \end{equation}
            Hence, minimizing the KL divergence between the variational family and the exact posterior distribution is equivalent to minimizing the variational objective  $\Psi(Q, \Pi, \bX_n)$ or maximizing the ELBO. 
            
            When the variational family has certain simple structure, in particular, the so-called \emph{mean field class}, there are efficient computational algorithms for finding $\hat{Q}$,  based on the well-known \emph{CAVI (coordinate ascent variational inference)} algorithm \cite{Jordan99, winn2005} which guarantees convergence to a local minimizer \cite{Blei17}.  Let $\theta=(\theta_1,\ldots, \theta_d) \in \Theta$ be a $d$-dimensional parameter, with $d$ potentially large.
            The mean-field class imposes posterior independence as: 
            \begin{equation}
            Q(\theta_1,\ldots, \theta_d)=\prod_{j=1}^d Q_j(\theta_j),
            \end{equation}  
         where $Q_j$ is a distribution for $\theta_j$.  By taking the derivative of the ELBO with respect to each of the $Q_j(\theta_j)$, one can arrive at the following coordinate ascent update: 
\begin{align}
\label{eq-cavi}
\hQ_j(\theta_j)&\propto \exp\left(E_{Q_{-j}}\left[\log p(\theta_j | \theta_{-j}, \bX_n\right]\right)\nonumber\\
&\propto \exp\left(E_{Q_{-j}}\left[\log p(\theta_j,  \theta_{-j}, \bX_n\right]\right)
\end{align}
         where $\theta_{-j}=(\theta_1,\ldots, \theta_{j-1}, \theta_{j+1},\ldots,\theta_n),$  and the expectation $E_{Q_{-j}}$ is taken with respect to all variational distributions but that of the $j$th component.   
        CAVI iteratively updates each coordinate by first initializing $Q_j(\theta_j)$ and then updating the variational distribution of each coordinate conditioned on the others according to  \eqref{eq-cavi}.
            
            When a statistical model has latent structures such as finite mixture models, topic models and stochastic block models, the dimension of  latent variables is typically of the same order as the sample size.    The CAVI algorithm  is not very efficient for large data sets as it requires  sweeping through the whole data set before updating the variational parameters at each iteration.  \emph{Stochastic variational inference (SVI)}  \cite{Hoffman13} is a popular alternative in this setting.  SVI employs  stochastic gradient  descent by computing the gradient of the ELBO based on mini batches.  

            \paragraph*{\textbf{Beyond the mean-field class}}
            CAVI and SVI critically depend on the mean-field assumption, 
            with this assumption ruling out posterior dependence across parameters and leading to under-estimation of posterior uncertainty.  This motivates more complex variational families, which tend to require tailored algorithms. \emph{Black-box VI  (BBVI)} algorithms \cite{Ranganath14}, including gradient based black-box VI,  have emerged as a popular class of such algorithms. \cite{geng} propose to utilize stochastic natural gradients within black-box VI to improve efficiency and address the common problem of large variance of gradient estimates.
            
            \paragraph*{\textbf{Amortized VB.}}  In traditional variational inference, parameters need to be optimized for each latent variable, which can be computationally intensive.   
            Amortized VI decreases this cost by building a map from  data points to the VB family.  This map is typically modeled by a deep neural network trained on  a data subset. The local VB parameter for the latent variable is computed using the output of the DNN map; 
            this is ``amortized'' since past computation is used to simplify future computation. Let  $f_\eta:\mathcal{X}\rightarrow {\Phi}$ be a feedforward neural network with parameters $\eta$ from the observation space $\mathcal{X}$ to the parameter space $\Phi$ of the variational family.  
              For observation $x_i$, the corresponding latent variable $\theta_i$ has conditional distribution $q_{f_\eta(x_i)}(\theta_i)$. Amortized variational  Bayes finds $\eta$ through:
\begin{align}
    \eta^* = \text{arg min}_{\eta} \mathrm{D}_{\mathrm{KL}}\left(\prod_{i=1}^n Q_{f_\eta(x_i)}(\theta_i)\|\Pi(\theta_i\mid \bX_n)\right). \label{eq:vb_am}
\end{align}
        
        Although amortized VI is a general framework, the most  popular application is the \emph{variational auto-encoder (VAE)}. The target generative model for data $\bX$ is 
         $\bX = {\boldsymbol G}({\boldsymbol Z}) + \bepsilon$, with ${\boldsymbol Z}$ latent data having a known distribution, $\bepsilon$ an additive noise independent of $\boldsymbol Z$, and $\boldsymbol G$ parametrized by a deep neural network. VAEs are a popular alternative to GANs for training deep generative models.  In a VAE, there is an encoder network where the distribution $\Pi(\boldsymbol Z\mid \bX_n, \theta)$ is amortized by a neural network mapping the data points to the variational family. 
                

 \subsection{Theory of variational Bayes} 

In order to verify the frequentist optimality properties of Bayesian posteriors, it is common to study 
contraction rates, model selection consistency, and asymptotic normality (known as Bernstein von-Mises (BvM) theorems).  Under the variational  Bayes framework, statistical inference is  based on the variational posterior instead of the original posterior, so it is natural to study frequentist optimality of VB posteriors.

In the asymptotic regime, we assume data $\bX^{(n)}$ are generated from $\P_{\theta^\star}^{(n)}$ and $n\to\infty$. The variational posterior
            \begin{align*}
              \hQ_n\in\argmin_{Q\in \cQ} \Psi(Q, \Pi, \bX^{(n)}),
            \end{align*}
         is said to have the contraction rate $\epsilon_n$ if 
            \begin{equation}
\E_{\theta^\star}^{(n)}[\hQ_n(d(\theta,\theta^\star)\le A_n\epsilon_n]\to1
            \end{equation}
        as $n\to\infty$ for any diverging sequence $A_n\to\infty$.   If the contraction rate $\epsilon_n$ matches the \emph{minimax optimal rate}, we say that the  variational posterior distribution is optimal.
        
 Recent work \citep{alquier2020concentration, zhang2020convergence,yang2020alpha} provided theoretical conditions under which the variational posterior is optimal.  These conditions imply that when the model is appropriately complex and the prior is sufficiently diffuse, which are standard conditions for establishing posterior contraction rates for the original posterior \cite{ghosal2000convergence}, then together with an assumption on the variational gap, the variational posterior distribution also  has optimal contraction rates. The variational  gap condition assumes there is $Q\in\cQ$ such that
        \begin{align}
        \label{vb-gap}
            \int \mathrm{D}_{\mathrm{KL}}(\P_{\theta}^{(n)}\|\P_{\theta^\star}^{(n)})Q(\d\theta) + \mathrm{D}_{\mathrm{KL}}(Q\| \Pi)\lesssim n\epsilon_{n}^2.
        \end{align}
The left side of \eqref{vb-gap} is an upper bound on the variational gap $\mathrm{D}_{\mathrm{KL}}(\hat{Q}\| \Pi(\theta\mid \bX_n)$. This upper bound is verified by ensuring that each term on the left is of order $O(n\epsilon_n^2)$.   \cite{alquier2020concentration} formulate this variational gap condition as an extension of prior mass conditions. If one restricts the VB family to be in the same class as the prior and the parameters to lie in a neighborhood of the true parameter, this condition reduces to the standard prior mass condition. 

 In addition,  \cite{pati2018statistical} and  \cite{yang2020alpha} developed variational Bayes theoretic frameworks that can deal with latent variable models.  \cite{alquier2020concentration} investigated the contraction properties of variational fractional posteriors with the likelihood raised to a fractional power. There are several studies that derived contraction rates of variational posteriors for specific statistical models - for example, mixture models \cite{cherief2018consistency}, sparse (Gaussian) linear regression  \citep{ray2021variational, yang2020variational}, sparse logistic linear regression \citep{ray2020spike}, and sparse factor models \citep{ning2021spike}.

\begin{figure}
\begin{center}
 \begin{tikzpicture}[scale=0.8, transform shape, framed, on grid,
   	node distance = \midY cm, 
   	inner frame xsep = 2.5ex,
   	inner frame ysep = 3.0ex,
    	box/.style = {rectangle,color=lightgray, text = black, text width =2.0cm, align=center},
    	wbox/.style = {rectangle,color=lightgray, text = black, text width = 4.0cm, align=center},
	    boxFramed/.style = {box, draw,inner sep = 3pt},
  	] 		  
  	\def\smallY{0.25}
    \def\midY{2.25}
    \def\smallX{1.2}
  	\node [wbox] (center) {\large  $\cM=\{1,\dots, M\}$};
   	\node [boxFramed, below = of center, xshift = -3*\smallX cm] (m1) {\normalLetters{\Theta_1}{\Pi_1} } ;
   	\node [boxFramed, below = of center, xshift =  -\smallX cm] (m2) {\normalLetters{\Theta_2}{\Pi_2} } ;   
   	\node [below = of center,  xshift = \smallX cm] (dots)  { \Large $\cdots$} ;
   	\node [boxFramed, below = of center,  xshift =  3*\smallX cm] (mM) {\normalLetters{\Theta_M}{\Pi_M}} ;   
   	\node [above = of dots, yshift=-0.8*\smallX cm] { \Large $\cdots$};
    \draw[->] (center)-- node[above] {\large $\alpha_1$} (m1.north);
    \draw[->] (center)-- node[right] {\large $\alpha_2$} (m2.north);
    \draw[->] (center)-- node[above] {\large $\alpha_M$} (mM.north);
\end{tikzpicture} 
\end{center}
\caption{The hierarchical prior distribution}
\label{hprior}
\end{figure}
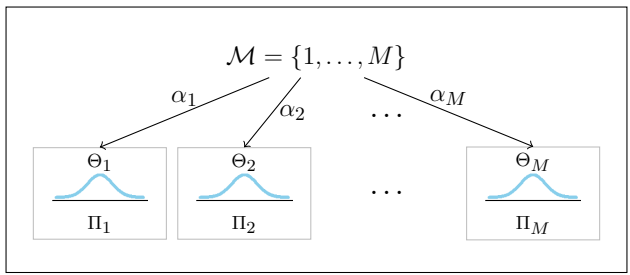

 \subsection{Adaptive Variational Bayes} 

 A notable recent development is a novel and general variational framework for adaptive statistical inference on a collection of model spaces
  \cite{ohn2021adaptive}.  The framework yields an \emph{adaptive variational posterior} that has optimal theoretical properties in terms of posterior contraction and model selection while enjoying  tractable computation. 
 
 In general, when performing  statistical inference the ``regularity'' of the true parameter is unknown and  adaptive inference aims to construct estimation procedures that are 
optimal with respect to the unknown true regularity. 
To do this,  one typically prepares \emph{multiple models} with different complexities, e.g. sparse linear regression models with different sparsity,  neural networks with different numbers of neurons or mixture models with different numbers of components, and then selects among them.
            To achieve adaptivity, frequentists usually  conduct (fully data-dependent) model selection before parameter estimation: e.g., via cross-validation or penalization.  There is some work on \emph{Bayesian adaptation} by imposing hierarchical priors on a collection of model spaces \cite{ghosal2008nonparametric}.
            
            Let $\cM$ denote a set of model indices and $\{\Theta_m\}_{m\in\cM}$ multiple disjoint (sub-)models with different complexities. Let $\Theta_{\cM}:=\cup_{m\in\cM}\Theta_m$ be an encompassing model. A (hierarchical) prior (illustrated by Figure \ref{hprior}) is given as 
            \begin{equation*}
                \Pi=\sum_{m\in\cM}\alpha_m\Pi_m,
            \end{equation*}
        where   $\alpha_m$ is the prior probability of model $\Theta_m$, $\sum_{m\in\cM}\alpha_m=1$, and   $\Pi_m$ is the prior distribution of $\theta$ within model $\Theta_m$.

The posterior distribution of $\Theta_{\cM}$ is 
            \begin{equation}
            \label{eq-hpost}
\Pi(\cdot|\bX_n)=\sum_{m\in\cM}\halpha_m\Pi(\cdot|\theta\in\Theta_m,\bX_n)
            \end{equation}
            where $\halpha_m=\Pi(\theta\in\Theta_m|\bX_n)$, which
            can be understood as a weighted average of the posteriors $\Pi(\cdot|\theta\in\Theta_m,\bX_n)
            $ on the individual models $(\Theta_m)_{m\in\cM}$.
        If the prior model probabilities $(\alpha_m)_{m\in\cM}$ are appropriately chosen, the posterior distribution on the encompassing model can be adaptively optimal \citep{lember2007universal,ghosal2008nonparametric,han2021oracle}.
        However, computing the posterior of $\Theta_{\cM}$ is challenging due to varying ``dimensions'' of the models 
        and the need to evaluate marginal likelihoods. 

           \cite{ohn2021adaptive} address these challenges via \emph{variational Bayes adaptation}. They approximate posterior \eqref{eq-hpost}  using a variational Bayes family over the encompassing model parameter space, using disjoint variational families $\{\cQ_m\}_{m\in\cM}$ over individual models with $\cQ_m\subset\cP(\Theta_m)$: 
    \begin{equation*}
        \cQ_{\cM}:=\left\{\sum_{m\in\cM}\gamma_mQ_m\mid  Q_m\in\cQ_m\right\}.
    \end{equation*}
They show that the variational posterior 
    \begin{equation*}
        \hQ_n\in\argmin_{Q\in\cQ_\cM}\Psi(Q, \Pi, \bX^{(n)})
    \end{equation*}
is of the form
 \begin{equation}
\hQ_n=\sum_{m\in\cM}\hgamma_{n,m}\hQ_{n,m}\in\cQ_{\cM}
        \end{equation}
for some `mixing weight' $(\hgamma_{n,m})_{m\in\cM}$ and `mixture components' $\hQ_{n,m}\in\cQ_m$ for $m\in\cM$.  The adaptive variational Bayes framework is summarized in Algorithm \ref{alg:general}.

\begin{algorithm}[t]
\caption{Adaptive variational Bayes}
   \label{alg:general}
\underline{Input}: data $X^{(n)}$, prior $\Pi=\sum_{m\in\cM}\alpha_m\Pi_m$, variational families $\{\cQ_m\}_{m\in\cM}$.
\null
    \begin{itemize}
        \item For every $m\in\cM$, compute the variational posterior of the submodel $\Theta_m$:
            \begin{equation}
               \hQ_{n,m}\in \argmin_{Q\in\cQ_m}\Psi(Q,\Pi_m,X^{(n)}).
            \end{equation}
        \item Compute the ``optimal model weight'' as
            \begin{equation}
                \hgamma_{n,m}
\propto\underbrace{\alpha_m}_{\text{prior}}\times\underbrace{\exp(-\Psi(\hQ_{n,m},\Pi_m,\bX^{(n)}))}_{\text{goodness of fit of $\hQ_{n,m}$}}
            \end{equation}
            for $m\in\cM$
        \end{itemize}
    
\null
\underline{Return}:
The adaptive variational posterior
            \begin{equation}
\hQ_n=\sum_{m\in\cM}\hgamma_{n,m}\hQ_{n,m}.
            \end{equation}
            \end{algorithm}

Computation of the adaptive variational posterior reduces to computing variational approximations for each individual model.  The framework is general and can be applied for adaptive inference in many statistical models where multiple submodels  of different complexities are available. The adaptive variational posterior has optimal contraction rates and strong model selection consistency when the true model is in $\mathcal{M}$. This theory has been applied to show optimal contraction for a rich variety of models, including finite mixtures, sparse factor models, deep neural networks and stochastic block models.

\subsection{Open questions and future directions}

\paragraph*{\textbf{Uncertainty quantification of the VB posterior}} It is well-known that variational posteriors tend to underestimate uncertainty of the posterior, so a central open question is how one can construct computationally efficient VB posteriors producing (a) credible balls with valid frequentist coverage and/or (b) posterior covariance matching that of the true posterior.


There is limited work on theory for statistical inference using the variational posterior, including credible intervals and hypothesis testing. For this, we need theorems to reveal a limiting distribution of the variational posterior as the sample size goes to infinity, just as the Bernstein-von Mises (BvM) theorem guarantees that the original posterior distribution converges to a Gaussian distribution under certain regularity conditions. An initial promising result along these lines is \cite{Wang18}, 
but there is substantial need for new research for broad classes of models and corresponding variational families.

\paragraph*{\textbf{Theoretical guarantees of gradient-based algorithms}} Existing theoretical guarantees for VB only apply to the global solution of the variational optimization problem. In practice, this optimization problem tends to be highly non-convex and algorithms are only guaranteed to converge to local optima. For certain variational families and model classes, these local optima can be dramatically different, so that there is a large sensitivity to the starting point of the algorithm. It is of critical importance to obtain guarantees on the algorithms being used and not just on inaccessible global optima. For example, can one obtain general theoretical guarantees for gradient-based black-box variational inference with or without warm-start conditions?

There is a parallel and growing literature on nonconvex optimization in other contexts, including providing reassurance that local optima can be sufficiently close in some cases \citep{mei2018landscape,foster2018uniform,loh2013regularized,li2019generalization,nikolakakis2022beyond}. However, to the best of our knowledge, there is no such work on theoretical aspects of local optima produced by variational Bayes.

\paragraph*{\textbf{VB based on generative models}} Richer variational families can be constructed using deep generative models such as normalizing flows \cite{rezende2015variational,liang2022fat}. Due to their impressive flexibility, the resulting variational posterior can approximate a very wide class of target posteriors accurately. Despite its practical usefulness and strong empirical performance, there is no theoretical support for such approaches - for example, providing upper bounds on the variational approximation gap or concentration properties. Choosing the 
neural network architecture and algorithmic tuning parameters involved in training to maximize computational efficiency and accuracy of posterior approximation is an additional important related area that may benefit from better theoretical understanding.

\paragraph*{\textbf{Online variational inference}} Given a prior distribution on an unknown parameter, the posterior distribution can be understood as an updated belief after observing the data. The updated posterior distribution can be used as a new prior distribution when new data arrive. The process can be repeated many times for analyzing streaming data \citep{minka, Yedidia_2016, kimY, JEONG2023107626}. At each step, the VB posterior can be used as a new prior instead of the original one for computational convenience \citep{linD, Loo2020GeneralizedVC, v.2018variational}.  It would be intriguing to investigate the statistical properties of the sequentially updated VB posterior.

\section{Discussion}
Tools for Bayesian computation are evolving at a rapid pace, thanks largely to recent developments in machine learning. We highlighted this phenomenon with four vignettes. The first vignette discussed sampling with the aid of generative models, particularly normalizing flows. The next two vignettes discussed different methods for handling the large $N$ regime.
Coresets take a variational approach to data compression, with recent methods leveraging deep neural networks to build flexible surrogate families; federated Bayesian learning methods instead distribute posterior computation over many computers. Finally, we covered  variational inference, which replaces the posterior with a tractable approximation. Many more vignettes could be written on similar topics, such as accelerating sampling with diffusion based generative models or accelerating approximate Bayesian computation using deep neural networks for data compression. We close with three themes, applicable to all vignettes, that we believe should receive future attention: accelerating inference using previous calculations, improving accessibility with new software, and providing theoretical support for empirically promising algorithms.

The status-quo in Bayesian computation is to start from scratch in each posterior inference problem, such as re-computing coresets after changing the prior, or estimating a new variaitonal approximation when applying an old model to new data. This is inefficient, as posterior inference in similar models must be somewhat informative about posterior inference in the current model. If the two models under consideration are directly comparable, such as posteriors under slightly different priors, then it may be easy to leverage previous calculations, e.g., by using warm starts in optimization routines. Problems arise when the two models have different dimensions, such a hierarchical models with an extra layer of parameters. We are hopeful that methods for similar problems in machine learning -- particularly transfer learning -- will play a role in developing general solutions for Bayesians.

Another common theme was the need for improved automation and accessibility. Implementing methods involving neural networks or other machine learning techniques in a robust and reliable fashion is a nontrivial task, often requiring significant time and expert knowledge. Given the breakneck speed at which machine learning progresses, careful implementations can be outdated before they have a chance for widespread adoption. The focus should be on developing software which is modular enough to withstand the next machine learning revolution, as well as user-friendly enough to be applied en-masse.

Finally, statisticians should be cautious with wholesale adoption of methods that achieve excellent practical performance at the expense of  theoretical guarantees. Fast ``approximations'' to posterior distributions that can be arbitrarily far from the exact posterior may be useful for black box prediction but fall far short of what is needed for reliable and reproducible Bayesian inferences. This is particularly key in scientific and policy applications in which one needs to appropriately characterize uncertainty in learning from data, acknowledging complexities that arise in practice such as model uncertainty, data contamination etc.
Guarantees are necessary to avoid highly misleading inferences and potentially catastrophic conclusions from the types of large and complex datasets that are being generated routinely in the sciences.

\bibliographystyle{imsart-nameyear.bst} 
\bibliography{combined_refs.bib}     


\end{document}